\documentclass[11pt]{article}

\usepackage{acl}

\usepackage{times}
\usepackage{latexsym}
\usepackage[T1]{fontenc}
\usepackage[utf8]{inputenc}
\usepackage{microtype}
\usepackage{inconsolata}

\usepackage{graphicx}
\usepackage{booktabs}
\usepackage{multirow}
\usepackage{makecell}
\usepackage[table]{xcolor}
\usepackage{tabularx}

\usepackage{amsmath}
\usepackage{algorithm}
\usepackage{algpseudocode}

\usepackage{url}
\usepackage{hyperref}

\usepackage[most]{tcolorbox}
\usepackage{xcolor}
\usepackage{listings}
\tcbuselibrary{listings,skins,breakable}
\usepackage[section]{placeins}

\usepackage{enumitem}
\setlist[itemize]{itemsep=2pt, topsep=2pt, parsep=0pt, partopsep=0pt}
\setlist[enumerate]{itemsep=2pt, topsep=2pt, parsep=0pt, partopsep=0pt}
\usepackage{makecell}
\newcommand{\mname}[2]{\makecell[l]{\textbf{#1}\\\textbf{#2}}}

\title{MemWeaver: Weaving Hybrid Memories for Traceable Long-Horizon Agentic Reasoning}

\author{
 \textbf{Juexiang Ye\textsuperscript{1}},
 \textbf{Xue Li\textsuperscript{1}},
 \textbf{Xinyu Yang\textsuperscript{1}},
 \textbf{Chengkai Huang\textsuperscript{2,3}},
 \textbf{Lanshun Nie\textsuperscript{1}},
 \textbf{Lina Yao\textsuperscript{2,4}},
 \textbf{Dechen Zhan\textsuperscript{1}}
\\
\\
 \textsuperscript{1}Harbin Institute of Technology,
 \textsuperscript{2}The University of New South Wales, 
\\
 \textsuperscript{3}Macquarie University,
 \textsuperscript{4}CSIRO's Data61
\\
 \small{\textbf{Correspondence:} \href{mailto:lixuecs@hit.edu.cn}{lixuecs@hit.edu.cn}}
}

\begin{document}
\maketitle

\begin{abstract}
Large language model-based agents operating in long-horizon interactions require memory systems that support temporal consistency, multi-hop reasoning, and evidence-grounded reuse across sessions. Existing approaches largely rely on unstructured retrieval or coarse abstractions, which often lead to temporal conflicts, brittle reasoning, and limited traceability. We propose MemWeaver, a unified memory framework that consolidates long-term agent experiences into three interconnected components: a temporally grounded graph memory for structured relational reasoning, an experience memory that abstracts recurring interaction patterns from repeated observations, and a passage memory that preserves original textual evidence. MemWeaver employs a dual-channel retrieval strategy that jointly retrieves structured knowledge and supporting evidence to construct compact yet information-dense contexts for reasoning. Experiments on the LoCoMo benchmark demonstrate that MemWeaver substantially improves multi-hop and temporal reasoning accuracy while reducing input context length by over 95\% compared to long-context baselines. 
Our data and code are available online.\footnote{\url{https://github.com/Chengkai-Huang/MemWeaver_code}}
\end{abstract}
    
\section{Introduction}
Large language model-based agents are increasingly deployed in long-horizon interactive settings, such as conversational assistants and personalized systems that span multiple sessions~\cite{park2023generative,maharana2024evaluating}. These scenarios require agents to maintain temporal consistency, accumulate user-specific knowledge, and reason over distant past interactions. However, limited context windows make reliance on internal representations alone infeasible, rendering external memory mechanisms essential~\cite{lewis2020retrieval,lee2024human}.

\begin{figure}[t]
  \centering
  \includegraphics[width=\linewidth]{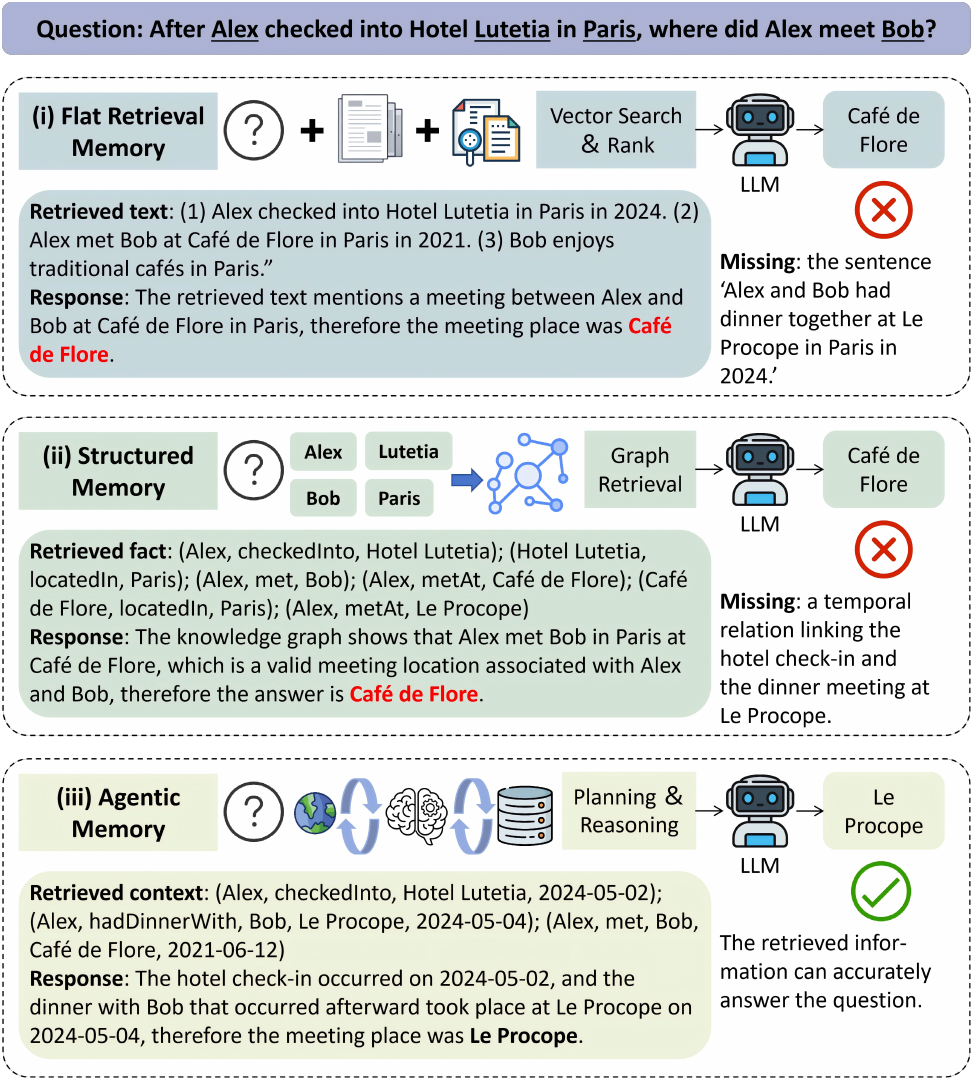}
  \caption{Comparison of flat retrieval memory, structured memory, and agentic memory from a reasoning perspective. While flat and structured memories can store and retrieve factual information, they do not explicitly model temporal relations between events, which limits their ability to answer temporally constrained queries. Agentic memory addresses this limitation by maintaining temporally grounded events and reasoning over their order.}
  \label{fig:intro}
\end{figure}


Recent LLM memory research has focused on retrieval-based augmentation, storing past interactions as passages or embeddings for semantic retrieval~\cite{zhong2024memorybank,borgeaud2022improving,jiao2026prunerag,li2021self}. Structured approaches include knowledge graphs for compositional reasoning~\cite{edge2024local,jimenez2024hipporag} and summary-based memories for preferences or strategies~\cite{park2023generative,shinn2023reflexion,huang2025pluralistic}. 



Despite progress, as illustrated in Figure \ref{fig:intro}, existing memory systems exhibit key limitations in long-term settings: flat retrieval memories lack relational and temporal structure, making them brittle for multi-hop and time-constrained queries~\cite{yang2018hotpotqa}; structured memories suffer from noisy extractions and accumulated conflicts due to missing session-level verification~\cite{zhong2023comprehensive}; and high-level abstractions are often weakly grounded, hindering traceability and correction~\cite{hong2023metagpt,yao2022react}. Most approaches treat memory as a passive retrieval buffer, overlooking systematic consolidation and update as interactions accumulate.

These limitations suggest that effective long-term agent memory requires a shift in perspective. Rather than viewing memory as a static repository, memory should be modeled as a \emph{consolidation process} that transforms episodic interaction traces into structured, reusable, and verifiable knowledge representations~\cite{xu2025mem}. This leads to a central research question:

\begin{quote}
\emph{How can an agent systematically consolidate long-term interaction histories into memory representations that support temporal consistency, compositional reasoning, cross-session generalization, and evidence-grounded decision making?}
\end{quote}

Addressing this question requires jointly designing \textbf{memory structures} and the \textbf{mechanisms} by which they are written, reviewed, and retrieved.
To this end, we propose \textbf{MemWeaver}, a consolidation-centric \textbf{tri-layer memory framework} for long-horizon agents. MemWeaver organizes long-term memory into three complementary layers: a \emph{temporally grounded knowledge graph} with normalized absolute-time metadata for \emph{time-aware multi-hop reasoning} and session-level conflict reconciliation; an \emph{experience abstraction} layer that clusters episodic windows and distills reusable experience items only when supported by consistent evidence across interactions; and a \emph{passage grounding} layer that preserves raw textual traces and links them to entities and experiences for \textbf{traceability}.

During inference, MemWeaver adopts a \textbf{dual-channel retrieval strategy}. Structured retrieval over the knowledge graph provides precise and compositional context, while evidence retrieval assembles tightly linked passages and experience items, complemented by a small number of globally retrieved passages for recall. The fused context enables the language model to generate responses that are both \emph{structurally informed} and \emph{evidence grounded}.

Our contributions are summarized as follows:

\begin{itemize}
    \item We formalize long-term agent memory from a reasoning-oriented perspective, highlighting the need for temporal grounding, compositional structure, and evidence traceability to support long-horizon inference.
    \item We propose MemWeaver, a consolidation-centric tri-layer memory framework that integrates temporally grounded knowledge graphs, experience abstraction, and evidence-linked passage memory through explicit writing and update mechanisms.
    \item We empirically demonstrate that MemWeaver substantially improves long-horizon reasoning performance, particularly on temporal and multi-hop tasks, while maintaining strong evidence traceability.
\end{itemize}

\section{Related Work}

\subsection{Memory Systems for LLM Agents}



LLM agents operating in long-horizon, multi-step settings must retain information beyond fixed context windows, motivating external memory mechanisms \cite{liu2023think,packer2023memgpt,huang2025survey,huang2025listwise,liu2021semi}. Existing approaches span retrieval-oriented memories that store interaction traces in external stores (e.g., vector databases) for cross-session access \cite{zhong2024memorybank,park2023generative}, structured or controlled architectures that regulate long-term information via hierarchies or controllers (e.g., MemGPT, SCM) \cite{packer2023memgpt,wang2023scm}, and abstraction-based memories that distill reusable reasoning patterns or task-level insights \cite{ouyang2025reasoningbank,zhang2025g}. More recent agentic frameworks further enable continual memory evolution by adding, revising, or deleting entries across sessions \cite{xu2025mem,chhikara2025mem0}. Despite this progress, many systems still rely on coarse, LLM-centric updates and provide limited support for temporally constrained reasoning and evidence-grounded correction at scale \cite{zhong2024memorybank}.

\subsection{Knowledge Graph–based Knowledge Organization}


Knowledge graphs (KGs) explicitly model entities and relations, and are increasingly used to structure knowledge for LLMs. Prior work leverages LLMs to construct or enrich KGs from unstructured text, or to instantiate task-specific graphs at inference time \cite{wu2025improving}. KGs have also been used to organize retrieved evidence and support multi-hop reasoning, including graph-guided inference frameworks such as Think-on-Graph and GraphRAG \cite{sun2023think,edge2024local}, as well as hybrid retrieval methods that combine vector similarity with graph-based exploration \cite{jimenez2024hipporag,gutierrez2025rag,yasunaga2021qa}. However, many KG-assisted pipelines introduce additional overhead and rely on task-specific graph construction and retrieval heuristics, and they typically lack explicit temporal grounding and continual updates, limiting their use in long-term interactive settings.


Despite the effectiveness of vector-similarity and LLM-assisted retrieval, response quality still depends heavily on how memories are constructed \cite{huang2025embedding}. We therefore propose MemWeaver, which unifies graph-structured knowledge, experience summaries, and passage-level evidence with an effective retrieval mechanism for long-horizon agentic reasoning.

\section{Problem Formulation}\label{sec:problem}

We consider a long-running conversational agent whose interaction history arrives incrementally.
The dialogue is modeled as a sequence of dialogue units (QA turns) with metadata:
\begin{equation}
x_i = \langle q_i, a_i, s_i, t_i \rangle,
\end{equation}
where $q_i$ and $a_i$ are the question and answer texts, $s_i$ is the speaker, and $t_i$ is the timestamp.
We use a unified encoder $\phi(\cdot)$ to obtain a semantic embedding:
\begin{equation}
e_i = \phi(\mathrm{text}(x_i)), \quad \mathrm{text}(x_i)=q_i \oplus a_i,
\end{equation}
which is used for clustering, routing, and retrieval. We model the agent memory state as a tri-layer set:
\begin{equation}
M = \{G, E, P\},
\end{equation}
where $G$ is Graph Memory (a knowledge graph of entities/relations with temporal or conditional attributes),
$E$ is Experience Memory (induced reusable experience items),
and $P$ is Passage Memory (a dense index over original text spans for evidence recall). The system supports two operations:
(i) \textbf{Memory writing} updates the state with a new unit:
\begin{equation}
M_{i+1} = \mathrm{Update}(M_i, x_i),
\end{equation}
and (ii) \textbf{Memory-based reasoning} retrieves a compact context for a query $Q$ and generates an answer:
\begin{equation}
y=\mathrm{LLM}(Q, C_{\mathrm{KG}}, C_{\mathrm{TXT}} \mid P_{\text{ans}}),
\end{equation}
where $(C_{\mathrm{KG}}, C_{\mathrm{TXT}})$ are retrieved from $M_i$.
Our goal is to maintain a high-quality $M$ over long dialogues, such that answers are accurate and supported by retrievable evidence.

\section{Methodology}

\begin{figure*}[t]
  \centering
  \includegraphics[width=0.85\linewidth]{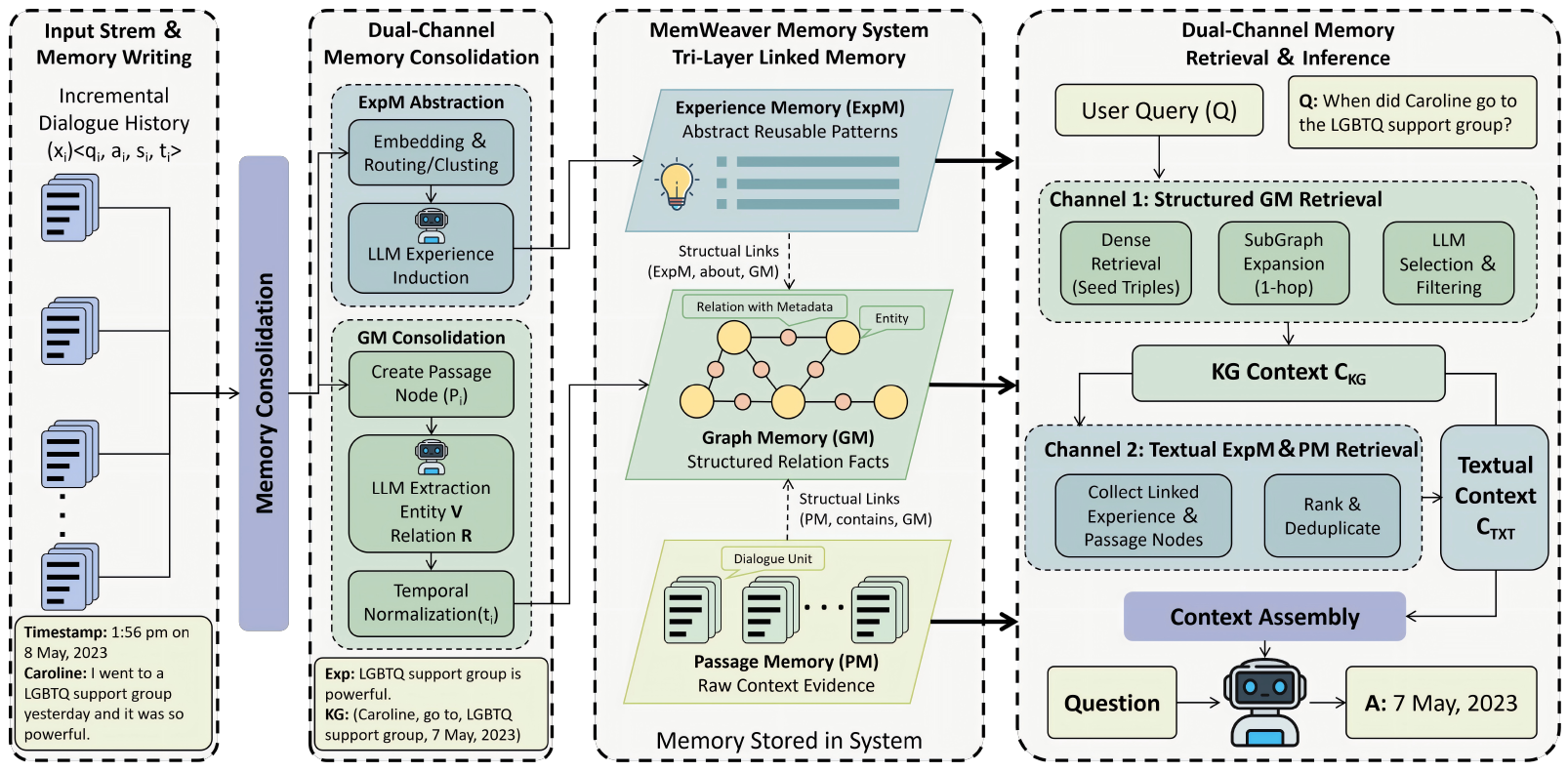}
  \caption{
  Overview of the MemWeaver framework.
  MemWeaver integrates Graph Memory (GM), Experience Memory (ExpM), and Passage Memory (PM)
  into a unified long-term memory system.
  Dialogue interactions are incrementally written into structured and unstructured memory,
  while dual-channel retrieval combines relational facts and textual evidence to construct a compact reasoning context for answer generation.
  }\label{fig:method}
\end{figure*}

This section presents \textbf{MemWeaver}, a tri-layer long-term memory system that integrates Graph Memory (GM), Experience Memory (ExpM), and Passage Memory (PM). Each component plays a complementary role during reasoning: GM supports compositional relational facts, ExpM captures reusable abstractions distilled from repeated interactions, and PM preserves verbatim evidence for robust recall.

\subsection{Tri-layer Memory Consolidation Framework}

MemWeaver explicitly links its three memory components rather than treating them as isolated stores.
Raw passages and experience items are attached to entity nodes in Graph Memory, allowing retrieved facts to be traced back to supporting evidence and enabling joint use of structured reasoning and evidence grounding.

The three memory carriers are loosely coupled through shared semantic representations and explicit structural links, allowing different memory types to be accessed and combined in a coordinated manner during inference.

At a high level, Graph Memory provides the structural backbone for organizing entities and their relations, while Passage Memory and Experience Memory are anchored to this structure as evidence-bearing nodes, together forming a unified memory representation. We describe these components in Sections~\ref{sec:kg}, \ref{sec:experience}, and \ref{sec:retrieval}.

\subsection{Graph Memory: Relational Fact Consolidation}
\label{sec:kg}

Graph Memory is designed as a relational fact consolidation layer that organizes
scattered factual statements across long interaction histories into a coherent
and reusable structure.
Instead of treating dialogue turns as independent evidence units, this layer
explicitly consolidates repeated and temporally grounded facts into stable
entity–relation representations.
Such consolidation enables consistent access to factual knowledge over long
time spans and provides a structural foundation for compositional and temporal
reasoning, which is difficult to achieve through unstructured memory alone.

MemWeaver represents its Graph Memory as a directed knowledge graph
$G = (V, \mathcal{R})$.
During memory construction, the graph consists of \emph{entity nodes}
and \emph{semantic relation edges}, representing consolidated factual knowledge
expressed in the dialogue history.
At inference time, \emph{passage nodes} and \emph{experience nodes} are
temporarily attached to entity nodes via \emph{structural edges}
(e.g., \texttt{contains} and \texttt{about}) to provide supporting evidence.

Each semantic relation $r$ is represented as a triple with metadata $m(r)$:
\begin{equation}
r=\langle h,\rho,u\rangle,\quad m(r)=\{\hat{t},c,\pi\},
\end{equation}
where $\hat{t}$ is a normalized absolute time expression, $c$ is an optional condition, $\pi$ is a provenance identifier for traceability and $m$ is metadata.

All LLM-based operations are performed using fixed prompt templates,
which are denoted as $P_{\text{ent}}$, $P_{\text{rel}}$, and $P_{\text{review}}$
for entity extraction, relation extraction, and session-level review, respectively.
Under this setting, given a newly observed dialogue unit $x_i$, the system incrementally writes it into the graph.
MemWeaver first creates a passage node that preserves the original
text together with its timestamp and speaker information.
It then performs a two-stage LLM-based extraction process,
where entities are identified in the first stage and relation candidates
are extracted under entity constraints in the second stage.
To improve temporal consistency, each candidate relation is further processed
by a dedicated normalization step that extracts an absolute time expression from the dialogue context:
\begin{equation}
\hat{t}=\eta(x_i,r),
\end{equation}

Where $\eta$ denotes the time-normalization function. After the initial write, MemWeaver performs a session-level review step.
All entities and relations introduced within the current session are serialized into a compact textual representation and verified by an LLM,
which outputs \texttt{add}, \texttt{update}, or \texttt{deny} operations to complete missing relations, correct erroneous ones, or remove noise. Finally, redundant semantic relations are removed, and the triple index $\mathcal{I}_T$ (a dense vector index over semantic relation triples) is rebuilt.
This index serves as the entry point for structured retrieval during inference. The overall procedure for KG writing and maintenance is summarized in Algorithm~\ref{alg:kg-build}.

\subsection{Experience Memory: Experience Abstraction}\label{sec:experience}

Experience Memory is designed to abstract reusable patterns from repeated interactions that are not well captured as isolated factual relations.
While Graph Memory consolidates explicit facts, many long-term signals such as preferences, tendencies, and recurring intents emerge only when multiple interactions are considered jointly.
By abstracting such recurring signals into experience-level representations, this layer supports cross-session generalization and complements fact-centric reasoning with higher-level behavioral knowledge.

Experience Memory abstracts historical dialogues by clustering semantically related units and inducing a compact set of reusable experience texts. Given the embedding of each dialogue unit, MemWeaver first applies DBSCAN clustering \cite{schubert2017dbscan,kulkarni2024survey} with cosine distance to obtain candidate topical groups. Since density-based clustering may still produce mixed-topic clusters, each candidate cluster is further validated by an LLM-based coherence check. Inconsistent clusters are returned to a pending queue for re-processing, whereas consistent ones are finalized as clusters and assigned a short theme summary $\texttt{center\_text}$, which serves as a semantic anchor for routing.

For each coherent cluster, MemWeaver induces a small set of experience items using an LLM. Each experience item is stored as a lightweight record consisting of an identifier, a semantic type, a content field, a list of supporting dialogue unit identifiers, and optional metadata. Each experience item must be explicitly supported by multiple dialogue units in the cluster and is stored together with its provenance information.
To reduce noise, MemWeaver applies filtering and deduplication heuristics to exclude small-talk patterns and remove near-duplicate items, ensuring that the resulting Experience Memory captures reusable signals rather than conversational artifacts.

In the online incremental setting, newly arriving dialogue units are routed to existing
clusters according to their similarity to cluster centers.
Let $e_i=\phi(\mathrm{text}(x_i))$ denote the embedding of a newly arriving dialogue unit, 
and let $\mu_j$ denote the center vector of cluster $j$:
\begin{equation}
\small{
\mu_j = \frac{1}{|C_j|} \sum_{x \in C_j} \phi(\mathrm{text}(x)),
j^\star = \arg\max_j \cos(e_i, \mu_j),
}
\end{equation}

MemWeaver performs three-way routing. If $\max_j \cos(e_i,\mu_j) \ge \tau_{\mathrm{high}}$, the dialogue unit is directly merged
into the best-matching cluster.
If $\tau_{\mathrm{low}} \le \max_j \cos(e_i,\mu_j) < \tau_{\mathrm{high}}$, the dialogue unit
is submitted to an LLM-based router for disambiguation among a shortlist of candidate clusters.
Formally, let $\mathcal{J}(x_i)$ denote the set of candidate clusters.
The router predicts the assignment as:
\begin{equation}
\small{
j^\star =
\mathrm{LLM}\!\left(
x_i,\ \{(\mathtt{center\_text}_j, \mathcal{S}_j)\}_{j \in \mathcal{J}(x_i)}
\mid P_{\mathrm{route}}
\right),
}
\end{equation}

where $\mathcal{S}_j$ denotes a small set of representative dialogue units sampled from cluster $j$. If the router returns \texttt{none}, the dialogue unit is appended to a pending buffer; Otherwise, it is assigned to cluster $j^\star$. If $\max_j \cos(e_i,\mu_j) < \tau_{\mathrm{low}}$, the dialogue unit is directly appended
to the pending buffer for future re-clustering once the buffer reaches a fixed window size.

To control update cost, each cluster maintains an $\texttt{add\_buffer}$ and only triggers an update when the buffer size exceeds a threshold, at which point the cluster center is recomputed and experience items are re-induced.
This strategy amortizes the expense of LLM-based extraction while maintaining
freshness under long-running interactions.
Algorithm~\ref{alg:exp-update} summarizes the complete induction and online update
procedure, including three-way routing, buffered updates, and reclustering.

\subsection{Dual-Channel Memory Retrieval}
\label{sec:retrieval}

MemWeaver constructs a dual-channel inference context by jointly retrieving structured relational facts and supporting textual evidence.
Structured retrieval over Graph Memory provides precise, compositional access to relations, while textual retrieval over Passage and Experience Memory improves recall and grounding.
Given a query $Q$, the retriever outputs a KG context $C_{\mathrm{KG}}$ and a textual context $C_{\mathrm{TXT}}$, which are fed to the backbone LM.
We denote by $k_r$, $k_p$, and $k_e$ the retrieval budgets for triples, passages, and experience items, respectively.

\textbf{Structured KG Retrieval.}
We embed semantic relation edges in Graph Memory (excluding structural links such as \texttt{contains} and \texttt{about}) and retrieve seed triples by cosine similarity.
MemWeaver then expands a bounded-hop neighborhood, filters candidates by similarity to control context size, and applies an LLM selector to choose the most useful triples.
To avoid overly narrow contexts, we augment the selected set with a few high-similarity triples for backfilling:
\begin{equation}
\small{
\begin{aligned}
\mathcal{R}^{\star}
&=\mathcal{R}_{\mathrm{LLM}}
\ \cup\ 
\operatorname{Top}_{k_r}\!\left(\mathcal{R}_{\mathrm{cand}};Q\right),
\end{aligned}}
\end{equation}
where $P_{\mathrm{select}}$ is a fixed prompt template for triple selection.

\textbf{Textual Evidence Retrieval.}
We collect evidence from two sources.
(i) For entities involved in $\mathcal{R}^\star$, we retrieve attached passage nodes via \texttt{contains} edges and experience nodes via \texttt{about} edges.
(ii) We additionally query a global dense retriever over all dialogue units to improve recall.
All retrieved texts are ranked by similarity to $Q$ and deduplicated by dialogue identifiers and content.

\textbf{Context Assembly.}
We serialize the selected subgraph $\mathcal{R}^\star$ to construct the graph context $C_{\mathrm{KG}}$, and aggregate the retrieved passages and experience items to form the textual context $C_{\mathrm{TXT}}$.
The backbone LLM then produces the final prediction conditioned on the assembled memory contexts:
\begin{equation}
    \hat{y} = f_{\mathrm{LLM}}\!\left(Q,\, C_{\mathrm{KG}},\, C_{\mathrm{TXT}}\right).
\end{equation}


\section{Experiments}
\begin{table*}[t]
\centering
\scriptsize
\setlength{\tabcolsep}{3.2pt}
\renewcommand{\arraystretch}{1.05}

\newcommand{\g}[1]{\cellcolor{gray!20}{#1}}
\resizebox{\textwidth}{!}{%
\begin{tabular}{l l|c c c c c c c c c c c c|c c c|c}
\toprule
\multirow{3}{*}{\textbf{Model}} & \multirow{3}{*}{\textbf{Method}}
& \multicolumn{12}{c|}{\textbf{Category}} & \multicolumn{4}{c}{\textbf{Overall}} \\
\cmidrule(lr){3-14}\cmidrule(lr){15-18}
& &
\multicolumn{3}{c}{\textbf{Multi-Hop}} &
\multicolumn{3}{c}{\textbf{Temporal}} &
\multicolumn{3}{c}{\textbf{Open-Domain}} &
\multicolumn{3}{c|}{\textbf{Single-Hop}} &
\multicolumn{3}{c|}{\textbf{Ranking}} &
\multirow{2}{*}{\textbf{Tokens}} \\
& &
\textbf{F1} & \textbf{BLEU} & \textbf{RGE-2} &
\textbf{F1} & \textbf{BLEU} & \textbf{RGE-2} &
\textbf{F1} & \textbf{BLEU} & \textbf{RGE-2} &
\textbf{F1} & \textbf{BLEU} & \textbf{RGE-2} &
\textbf{F1} & \textbf{BLEU} & \textbf{RGE-2} & \\
\midrule

\multirow{5}{*}{\mname{GPT}{-4o-mini}}
& LoCoMo
& 24.35 & 16.91 & 8.50
& 22.54 & 16.93 & 9.03
& 15.39 & 13.59 & 3.86
& \textbf{42.39} & 31.65 & \textbf{28.29}
& 2.00 & 2.25 & 2.25
& 21,625 \\
& MemoryBank
& 6.10 & 4.52 & 1.57
& 5.04 & 3.02 & 0.56
& 6.32 & 3.27 & 1.04
& 8.42 & 3.96 & 3.21
& 4.75 & 5.00 & 4.50
& 1,513 \\
& ReadAgent
& 9.15 & 6.48 & 2.47
& 12.60 & 8.87 & 0.95
& 5.31 & 5.12 & 0.55
& 9.67 & 7.66 & 2.99
& 4.25 & 4.00 & 4.50
& 643 \\
& A-Mem
& 23.68 & 16.74 & 8.42
& 38.77 & 33.71 & 15.58
& 12.50 & 11.29 & 3.96
& 35.13 & 28.96 & 21.53
& 2.75 & 2.75 & 2.50
& 2,731 \\
& \g{\textbf{MemWeaver}}
& \g{\textbf{26.00}} & \g{\textbf{17.08}} & \g{\textbf{10.90}}
& \g{\textbf{50.83}} & \g{\textbf{43.72}} & \g{\textbf{23.36}}
& \g{\textbf{20.73}} & \g{\textbf{15.63}} & \g{\textbf{5.23}}
& \g{39.20} & \g{\textbf{33.58}} & \g{25.96}
& \g{\textbf{1.25}} & \g{\textbf{1.00}} & \g{\textbf{1.25}}
& \g{672} \\
\midrule

\multirow{5}{*}{\mname{Llama3.2}{-3B}}
& LoCoMo
& 8.09 & 6.88 & 2.29
& 5.00 & 5.45 & 0.77
& 7.08 & 5.83 & 0.65
& 11.92 & 8.63 & 5.02
& 2.75 & 2.75 & 4.00
& 22,312 \\
& MemoryBank
& 5.04 & 3.34 & 1.15
& 2.06 & 1.19 & 0.13
& 4.52 & 2.09 & 0.83
& 6.78 & 2.90 & 2.34
& 4.50 & 4.50 & 4.75
& 1,553 \\
& ReadAgent
& 2.47 & 1.78 & 2.47
& 3.01 & 3.01 & 3.01
& 5.57 & 5.22 & \textbf{5.07}
& 3.25 & 2.51 & 3.25
& 4.25 & 4.50 & 2.75
& 461 \\
& A-Mem
& 9.96 & \textbf{13.21} & 6.62
& 8.62 & 6.76 & 4.41
& 5.03 & 5.78 & 0.87
& 19.85 & 19.70 & 16.98
& 2.50 & 2.00 & 2.25
& 2508 \\
& \g{\textbf{MemWeaver}}
& \g{\textbf{11.86}} & \g{13.07} & \g{\textbf{6.94}}
& \g{\textbf{11.37}} & \g{\textbf{13.16}} & \g{\textbf{8.67}}
& \g{\textbf{9.64}}  & \g{\textbf{7.77}}  & \g{2.11}
& \g{\textbf{21.22}} & \g{\textbf{22.27}} & \g{\textbf{20.28}}
& \g{\textbf{1.00}} & \g{\textbf{1.25}} & \g{\textbf{1.25}}
& \g{1000} \\
\midrule

\multirow{5}{*}{\mname{Llama3.2}{-1B}}
& LoCoMo
& 10.37 & 8.10 & 2.85
& 15.51 & 12.83 & 1.01
& 11.94 & 10.64 & 2.20
& 13.90 & 10.64 & 4.88
& 3.25 & 2.75 & 3.25
& 22,312 \\
& MemoryBank
& 4.80 & 3.36 & 1.12
& 1.89 & 1.61 & 0.06
& 5.72 & 3.58 & 0.83
& 6.42 & 3.02 & 2.15
& 5.00 & 5.00 & 4.25
& 1,553 \\
& ReadAgent
& 5.96 & 5.12 & 0.53
& 1.93 & 2.30 & 0.00
& \textbf{12.46} & \textbf{11.17} & \textbf{5.47}
& 7.75 & 6.03 & 1.19
& 3.25 & 3.25 & 4.00
& 665 \\
& A-Mem
& 11.03 & 9.21 & 3.35
& 17.76 & 12.98 & 3.64
& 12.34 & 7.58 & 2.85
& 19.20 & 15.01 & 7.63
& 2.25 & 2.50 & 2.25
& 2647 \\
& \g{\textbf{MemWeaver}}
& \g{\textbf{12.96}} & \g{\textbf{9.77}}  & \g{\textbf{3.72}}
& \g{\textbf{24.11}} & \g{\textbf{17.42}} & \g{\textbf{7.53}}
& \g{12.41} & \g{9.94}  & \g{3.04}
& \g{\textbf{20.17}} & \g{\textbf{15.05}} & \g{\textbf{11.78}}
& \g{\textbf{1.25}} & \g{\textbf{1.50}} & \g{\textbf{1.25}}
& \g{1000} \\
\midrule

\multirow{5}{*}{\mname{Qwen2.5}{-1.5B}}
& LoCoMo
& 10.23 & 7.96 & 2.14
& 7.65 & 6.76 & 0.21
& 11.58 & 9.74 & 2.37
& 12.93 & 9.18 & 4.14
& 2.75 & 3.00 & 3.00
& 22,424 \\
& MemoryBank
& 6.63 & 5.02 & 1.77
& 3.43 & 2.85 & 0.20
& 5.80 & 3.40 & 0.82
& 9.41 & 4.85 & 3.38
& 4.25 & 4.50 & 4.00
& 1,555 \\
& ReadAgent
& 6.61 & 4.93 & 0
& 2.55 & 2.51 & 0
& 5.31 & 12.24 & 0
& 10.13 & 7.54 & 0
& 4.75 & 4.00 & 5.00
& 752 \\
& A-Mem
& 12.57 & 9.94 & 5.17
& 14.54 & 12.45 & 1.73
& 10.83 & 9.21 & 2.43
& 20.06 & 14.86 & 11.07
& 2.25 & 2.50 & 2.00
& 2,574 \\
& \g{\textbf{MemWeaver}}
& \g{\textbf{21.91}} & \g{\textbf{16.47}} & \g{\textbf{7.25}}
& \g{\textbf{46.07}} & \g{\textbf{38.47}} & \g{\textbf{17.74}}
& \g{\textbf{17.94}} & \g{\textbf{15.86}} & \g{\textbf{4.23}}
& \g{\textbf{32.80}} & \g{\textbf{27.42}} & \g{\textbf{18.52}}
& \g{\textbf{1.00}} & \g{\textbf{1.00}} & \g{\textbf{1.00}}
& \g{734} \\
\bottomrule

\end{tabular}%
}
\vspace{-1em}
\caption{Experimental results on the LoCoMo dataset across four question types (Multi-Hop, Temporal, Open-Domain, and Single-Hop). Results are reported in F1, BLEU-1 (\%), and RGE-2 (\%). RGE-2 denotes ROUGE-2. Best results within each backbone are in \textbf{bold}, and MemWeaver is highlighted in gray. Ranking indicates the average rank across categories (Rank 1 is best; lower is better), computed separately for F1, BLEU-1, and RGE-2. Tokens is the average number of input tokens per query.}
\label{tab:locomo}

\end{table*}

\subsection{Experiment Preparation}

\textbf{Datasets.} Following previous work \cite{maharana2024evaluating,zhong2024memorybank,lee2024human,xu2025mem}, we evaluate MemWeaver on the LoCoMo dataset \cite{maharana2024evaluating}, a benchmark designed for long-term conversational question answering with extended multi-session dialogue histories.
Compared with prior conversational datasets that contain around 1K tokens over a small number of sessions, LoCoMo features substantially longer conversations, averaging approximately 9K tokens and spanning up to 35 sessions. 
This setting makes LoCoMo particularly suitable for evaluating models’ ability to retrieve, integrate, and reason over long-range contextual information. LoCoMo includes five question types (Single-Hop, Multi-Hop, Temporal, Open-Domain, and Adversarial), and we focus on the four answerable categories in our main experiments. For detailed dataset statistics and category definitions, please refer to Appendix~\ref{app:impl}.

\textbf{Baselines.} For fair comparison, we evaluate MemWeaver against four representative baselines: LoCoMo \cite{maharana2024evaluating}, which relies on long-context prompting over the dialogue history;
MemoryBank \cite{zhong2024memorybank}, a retrieval-based conversational memory framework;
ReadAgent \cite{lee2024human}, a reading-based agent that selectively summarizes and retrieves dialogue history;
and A-Mem \cite{xu2025mem}, an agentic memory system based on atomic memory construction.
A detailed description of the baselines is provided in Appendix \ref{app:experiment}.

\textbf{Evaluation Metrics.} Following prior work on LoCoMo \cite{maharana2024evaluating}, we employ two primary evaluation metrics.
We report token-level F1 to assess answer accuracy by balancing precision and recall,
and BLEU-1 to measure lexical overlap between generated responses and ground-truth answers. In addition, we report the average number of input tokens per query,
which reflects the inference-time context length and associated computational cost.
We further report additional metrics, including exact match (EM), METEOR, ROUGE-L and SBERT similarity, in the Appendix \ref{app:experiment}.

\subsection{Implementation Details}

We evaluate four backbone LLMs: GPT-4o-mini \cite{hurst2024gpt}, Llama3.2-3B/1B \cite{grattafiori2024llama}, and Qwen2.5-1.5B \cite{yang2025qwen2}. All methods share identical system prompts and output formats. To decouple memory construction from inference-time reasoning, we use DeepSeek-V3.2 (non-thinking) \cite{deepseek2025deepseek} to build all MemWeaver memory components \emph{offline} (never at inference), since structured memory writing (entity/relation extraction, temporal normalization, and session-level verification) is unreliable for small LMs. At inference, MemWeaver retrieves $C_{\mathrm{KG}}$ and $C_{\mathrm{TXT}}$ via top-$k$ retrieval with default $k_r=k_p=k_e=6$ (category-specific adjustments when needed). We use \texttt{all-minilm-l6-v2} \cite{reimers2019sentence} for embeddings, and follow each baseline’s original settings while matching inference-time context length to MemWeaver. More details are provided in Appendix \ref{app:impl}.

\subsection{Main Results}

\begin{table}[H]
\centering
\caption{Comparison of memory usage and retrieval time across different memory methods.}
\label{tab:memory_time}
\footnotesize
\setlength{\tabcolsep}{4pt}
\renewcommand{\arraystretch}{1.12}

\begin{tabular}{l|ccc|c}
\hline
\multirow{2}{*}{\textbf{Method}} & \multicolumn{3}{c|}{\textbf{Memory Usage (MB)}} & \textbf{Retrieval} \\
\cline{2-4}
& \textbf{GM} & \textbf{ExpM} & \textbf{Total} & \textbf{Times (ms)}\\
\hline
MemoryBank & - & - & 7.23 & 17.07 $\pm$ 3.61 \\
A-Mem      & - & - & 18.29 & 16.00 $\pm$ 7.18 \\
MemWeaver  & 5.24 & 8.07 & 13.31 & 41.57 $\pm$ 12.85 \\
\hline
\end{tabular}
\end{table}

\begin{table*}[t]
\centering
\small
\setlength{\tabcolsep}{5pt}
\renewcommand{\arraystretch}{1.08}

\resizebox{0.93\textwidth}{!}{%
\begin{tabular}{l l|c c c|c c c|c c c|c c c}
\toprule
\multirow{2}{*}{\textbf{Model}} & \multirow{2}{*}{\textbf{Method}}
& \multicolumn{3}{c|}{\textbf{Multi-Hop}}
& \multicolumn{3}{c|}{\textbf{Temporal}}
& \multicolumn{3}{c|}{\textbf{Open-Domain}}
& \multicolumn{3}{c}{\textbf{Single-Hop}} \\
\cmidrule(lr){3-5}\cmidrule(lr){6-8}\cmidrule(lr){9-11}\cmidrule(lr){12-14}
& & \textbf{F1} & \textbf{BLEU-1} & \textbf{RGE-2}
  & \textbf{F1} & \textbf{BLEU-1} & \textbf{RGE-2}
  & \textbf{F1} & \textbf{BLEU-1} & \textbf{RGE-2}
  & \textbf{F1} & \textbf{BLEU-1} & \textbf{RGE-2} \\
\midrule

\multirow{3}{*}{\textbf{GPT-4o-mini}}
& w/o ExpM
& 24.86 & 15.70 & 9.83 
& 48.67 & 41.60 & 21.74 
& 19.65 & 15.31 & 5.05 
& 38.07 & 32.66 & 23.84 \\
& w/o GM
& 13.97 &  8.28 & 4.75 &  9.58 &  5.60 & 1.04 & 10.30 &  9.05 & 3.01 & 13.12 & 11.06 & 6.45 \\
\rowcolor{gray!20}
& \textbf{MemWeaver}
& \textbf{26.00} & \textbf{17.08} & \textbf{10.90} & \textbf{50.83} & \textbf{43.72} & \textbf{23.36} & \textbf{20.73} & \textbf{15.63} & \textbf{5.23} & \textbf{39.20} & \textbf{33.58} & \textbf{25.96} \\
\midrule

\multirow{3}{*}{\textbf{Qwen2.5-1.5B}}
& w/o ExpM
& 19.59 & 14.13 & 6.62 
& 45.30 & 37.75 & 15.39 
& 17.33 & 15.07 & 3.97 
& 32.38 & 26.74 & 16.85 \\
& w/o GM
& 13.48 &  9.85 & 4.31 &  5.78 &  4.88 & 0.83 & 10.33 &  9.76 & 1.61 & 13.84 & 11.46 & 6.35 \\
\rowcolor{gray!20}
& \textbf{MemWeaver}
& \textbf{21.91} & \textbf{16.47} & \textbf{7.25} & \textbf{46.07} & \textbf{38.47} & \textbf{17.74} & \textbf{17.94} & \textbf{15.86} & \textbf{4.23} & \textbf{32.80} & \textbf{27.42} & \textbf{18.52} \\
\bottomrule
\end{tabular}%
}
\vspace{-1em}
\caption{Ablation study of MemWeaver under two backbone models (GPT-4o-mini and Qwen2.5-1.5B). Results are reported in F1, BLEU-1, and RGE-2 (ROUGE-2) (\%).}
\label{tab:ablation}
\end{table*}

\textbf{Performance Analysis.}
Table~\ref{tab:locomo} reports category-wise results and overall ranking on LoCoMo under four backbone language models.
Ranking is the average rank across the four categories (lower is better), computed separately for F1, BLEU-1, and ROUGE-2.
Across backbones, MemWeaver achieves the best overall ranking in most settings, indicating consistently strong performance across diverse question types.

MemWeaver is particularly effective on \emph{Multi-Hop} and \emph{Temporal} questions that require cross-session reasoning. With GPT-4o-mini, MemWeaver improves Multi-Hop F1 from 24.35 (LoCoMo) / 23.68 (A-Mem) to 26.00, and boosts Temporal F1 from 38.77 (A-Mem) to 50.83.
Similar trends are observed for smaller backbones: for Qwen2.5-1.5B, MemWeaver improves Temporal F1 from 14.54 to 46.07.
We attribute the larger gains on smaller backbones to structured retrieval, which externalizes key factual and temporal cues and reduces reliance on parametric knowledge and long-context reasoning.

Compared with LoCoMo, MemWeaver achieves higher or comparable accuracy while using substantially shorter inference contexts.
MemoryBank performs poorly across most categories, suggesting that flat retrieval over unstructured memories is insufficient for complex long-horizon reasoning.
Relative to the agentic baseline A-Mem, MemWeaver yields consistent improvements (notably on \emph{Temporal} and \emph{Open-Domain}), highlighting the benefit of separating temporally grounded Graph Memory from experience-level abstractions.

\textbf{Token Efficiency.}
MemWeaver also substantially reduces inference-time context length.
LoCoMo typically requires over 22K input tokens per query due to long-context prompting, whereas MemWeaver stays within 1K tokens across all backbones, reducing input length by over 95\% while maintaining or improving performance.
Compared with A-Mem, MemWeaver further reduces context length by roughly 2--4$\times$.
Memory and latency are reported in Table~\ref{tab:memory_time}, showing that MemWeaver trades modest retrieval overhead for stronger accuracy and substantially shorter inputs, making it suitable for long-running and resource-constrained deployments.

\subsection{Ablation Study}
We conduct an ablation study to analyze the contribution of the two core components in MemWeaver: Graph Memory (GM) and Experience Memory (ExpM). We evaluate two variants, w/o GM and w/o ExpM, under two backbone models, focusing on answerable question types (Multi-Hop, Temporal, Open-Domain, and Single-Hop). In all settings, the global passage retrieval channel is preserved to ensure that observed performance differences mainly reflect the impact of structured and abstracted memory components rather than access to raw textual evidence.

As shown in Table~\ref{tab:ablation}, removing ExpM leads to consistent but moderate performance drops across tasks, indicating its role in capturing reusable abstractions that support cross-session reasoning. In contrast, removing GM causes severe degradation, especially on Multi-Hop and Temporal questions, highlighting the importance of graph-based structural and temporal organization for compositional reasoning. Overall, the full MemWeaver model achieves the best and most balanced performance, demonstrating that Graph Memory and Experience Memory play complementary roles: GM provides structured factual reasoning, while ExpM supplies higher-level abstraction, together enabling robust long-term conversational reasoning.

\subsection{Hyperparameter Analysis}

\begin{figure}[!htbp]
\centering

\begin{minipage}[t]{0.49\linewidth}
  \centering
  \includegraphics[width=\linewidth]{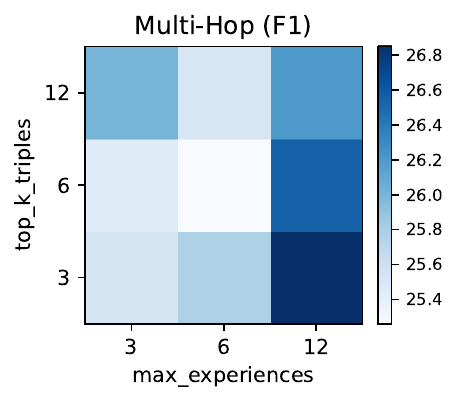}
  \vspace{-0.3em}
  {\small (a) Multi-Hop}
\end{minipage}\hfill
\begin{minipage}[t]{0.49\linewidth}
  \centering
  \includegraphics[width=\linewidth]{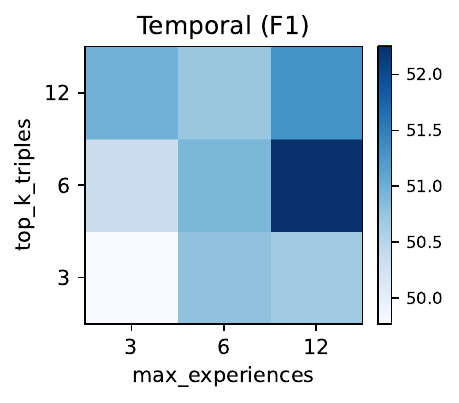}
  \vspace{-0.3em}
  {\small (b) Temporal}
\end{minipage}

\vspace{0.3em}

\begin{minipage}[t]{0.49\linewidth}
  \centering
  \includegraphics[width=\linewidth]{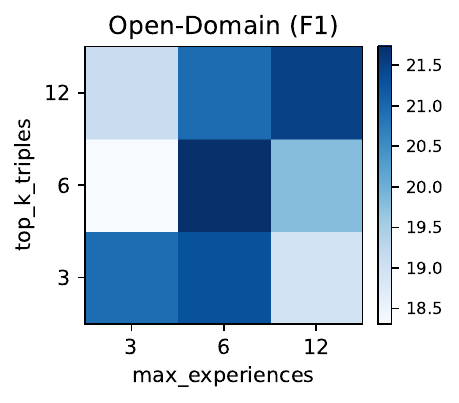}
  \vspace{-0.3em}
  {\small (c) Open-Domain}
\end{minipage}\hfill
\begin{minipage}[t]{0.49\linewidth}
  \centering
  \includegraphics[width=\linewidth]{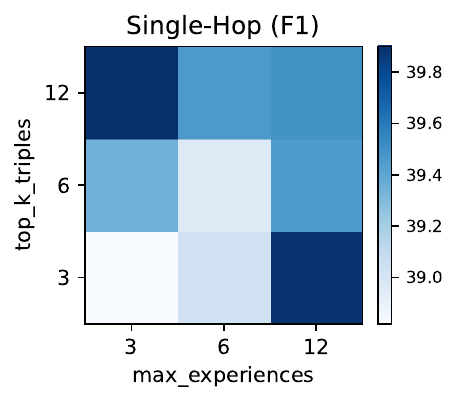}
  \vspace{-0.3em}
  {\small (d) Single-Hop}
\end{minipage}

\caption{Hyperparameter sensitivity analysis of MemWeaver.
Each heatmap reports performance under different retrieval configurations for (a) Multi-Hop, (b) Temporal, (c) Open-Domain, and (d) Single-Hop questions.}
\label{fig:hyperparam_heatmaps}
\end{figure}

We investigate MemWeaver's sensitivity to retrieval budgets, varying the number of retrieved graph triples and textual memory items. Figure~\ref{fig:hyperparam_heatmaps} reports heatmaps across the four answerable categories under different retrieval configurations.
Overall, MemWeaver is stable across a wide range of settings, with only mild performance variation and no abrupt degradation as retrieval sizes change.
Increasing the number of retrieved items does not consistently improve performance and often shows diminishing or negligible gains.
These results suggest that MemWeaver does not rely on large retrieval volumes; instead, its structured memory design enables efficient and selective retrieval, prioritizing relevance over quantity.

Taken together, the proposed memory system remains both stable and efficient, and its effectiveness stems from precise organization and targeted retrieval rather than brute-force context expansion.

\subsection{Human Evaluation}

\begin{figure}[!htbp]
  \centering
  \includegraphics[width=\columnwidth]{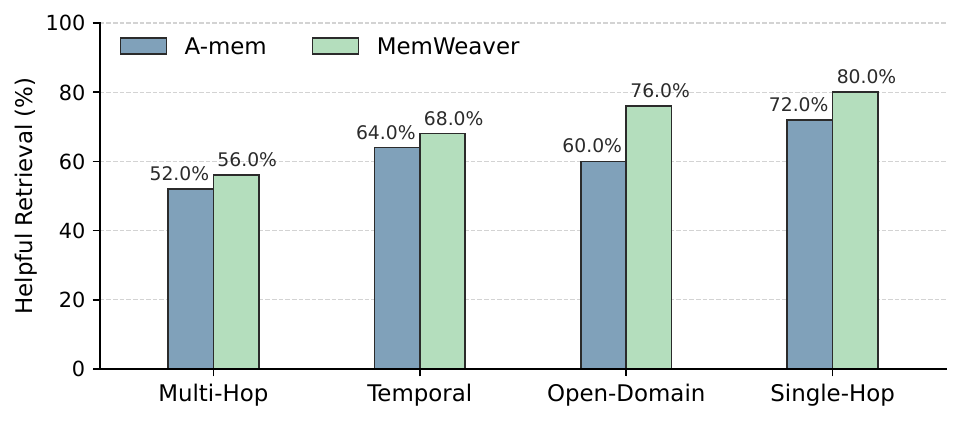}
  \vspace{-2em}
  \caption{
    Human evaluation of evidence support quality for A-Mem and MemWeaver across four question categories.
    Using GPT-4o-mini as the backbone, human annotators assess, on 25 sampled questions per category,
    whether the retrieved knowledge provides sufficient support for the generated answers.
  }
  \label{fig:human_eval}
\end{figure}

To complement automatic metrics, we conduct a human evaluation to assess whether the retrieved knowledge provides sufficient support for the generated answers.
As shown in Figure~\ref{fig:human_eval}, MemWeaver consistently outperforms A-Mem across all question categories, indicating more reliable and evidence-grounded retrieval. Details are provided in Appendix \ref{app:experiment:human}.

\section{Conclusion}

In this paper, we propose MemWeaver, a structured long-term memory system for LLM agents that unifies graph-structured knowledge, abstracted experiences, and passage-level evidence to support scalable and traceable conversational reasoning. MemWeaver incrementally maintains a temporally grounded knowledge graph with LLM-based verification, derives reusable experience items from clustered interactions, and retains a global passage retrieval channel for accessing original evidence. Experiments on the LoCoMo benchmark across multiple backbone models show its effectiveness and consistent improvements in long-horizon conversational question answering with efficient inference. 


\section*{Limitations}
Although MemWeaver achieves encouraging results, we recognize several directions for future exploration.
First, while the system is able to organize memory in a structured manner, the quality of memory organization may still be influenced by the inherent capabilities of the underlying language model. Different models may produce slightly different abstractions or associations when constructing memory.
Second, the current implementation primarily focuses on textual interactions. Extending the framework to incorporate multimodal information such as images or audio is a promising direction for future work.

\bibliography{custom}

\begin{thebibliography}{39}
\providecommand{\natexlab}[1]{#1}

\bibitem[{Borgeaud et~al.(2022)Borgeaud, Mensch, Hoffmann, Cai, Rutherford, Millican, van~den Driessche, Lespiau, Damoc, Clark, de~Las~Casas, Guy, Menick, Ring, Hennigan, Huang, Maggiore, Jones, Cassirer, Brock, Paganini, Irving, Vinyals, Osindero, Simonyan, Rae, Elsen, and Sifre}]{borgeaud2022improving}
Sebastian Borgeaud, Arthur Mensch, Jordan Hoffmann, Trevor Cai, Eliza Rutherford, Katie Millican, George van~den Driessche, Jean{-}Baptiste Lespiau, Bogdan Damoc, Aidan Clark, Diego de~Las~Casas, Aurelia Guy, Jacob Menick, Roman Ring, Tom Hennigan, Saffron Huang, Loren Maggiore, Chris Jones, Albin Cassirer, and 9 others. 2022.
\newblock \href {https://proceedings.mlr.press/v162/borgeaud22a.html} {Improving language models by retrieving from trillions of tokens}.
\newblock In \emph{International Conference on Machine Learning, {ICML} 2022, 17-23 July 2022, Baltimore, Maryland, {USA}}, volume 162 of \emph{Proceedings of Machine Learning Research}, pages 2206--2240. {PMLR}.

\bibitem[{Chhikara et~al.(2025)Chhikara, Khant, Aryan, Singh, and Yadav}]{chhikara2025mem0}
Prateek Chhikara, Dev Khant, Saket Aryan, Taranjeet Singh, and Deshraj Yadav. 2025.
\newblock \href {https://doi.org/10.48550/ARXIV.2504.19413} {Mem0: Building production-ready {AI} agents with scalable long-term memory}.
\newblock \emph{CoRR}, abs/2504.19413.

\bibitem[{DeepSeek-AI et~al.(2025)DeepSeek-AI, Mei, Lin, Xue, Wang, Xu, Wu, Zhang, Lin, Dong et~al.}]{deepseek2025deepseek}
Aixin~Liu DeepSeek-AI, Aoxue Mei, Bangcai Lin, Bing Xue, Bingxuan Wang, Bingzheng Xu, Bochao Wu, Bowei Zhang, Chaofan Lin, Chen Dong, and 1 others. 2025.
\newblock Deepseek-v3. 2: Pushing the frontier of open large language models.
\newblock \emph{arXiv preprint arXiv:2512.02556}.

\bibitem[{Edge et~al.(2024)Edge, Trinh, Cheng, Bradley, Chao, Mody, Truitt, and Larson}]{edge2024local}
Darren Edge, Ha~Trinh, Newman Cheng, Joshua Bradley, Alex Chao, Apurva Mody, Steven Truitt, and Jonathan Larson. 2024.
\newblock \href {https://doi.org/10.48550/ARXIV.2404.16130} {From local to global: {A} graph {RAG} approach to query-focused summarization}.
\newblock \emph{CoRR}, abs/2404.16130.

\bibitem[{Gutierrez et~al.(2024)Gutierrez, Shu, Gu, Yasunaga, and Su}]{jimenez2024hipporag}
Bernal~Jimenez Gutierrez, Yiheng Shu, Yu~Gu, Michihiro Yasunaga, and Yu~Su. 2024.
\newblock \href {http://papers.nips.cc/paper\_files/paper/2024/hash/6ddc001d07ca4f319af96a3024f6dbd1-Abstract-Conference.html} {Hipporag: Neurobiologically inspired long-term memory for large language models}.
\newblock In \emph{Advances in Neural Information Processing Systems 38: Annual Conference on Neural Information Processing Systems 2024, NeurIPS 2024, Vancouver, BC, Canada, December 10 - 15, 2024}.

\bibitem[{Guti{\'{e}}rrez et~al.(2025)Guti{\'{e}}rrez, Shu, Qi, Zhou, and Su}]{gutierrez2025rag}
Bernal~Jim{\'{e}}nez Guti{\'{e}}rrez, Yiheng Shu, Weijian Qi, Sizhe Zhou, and Yu~Su. 2025.
\newblock \href {https://openreview.net/forum?id=LWH8yn4HS2} {From {RAG} to memory: Non-parametric continual learning for large language models}.
\newblock In \emph{Forty-second International Conference on Machine Learning, {ICML} 2025, Vancouver, BC, Canada, July 13-19, 2025}. OpenReview.net.

\bibitem[{Hong et~al.(2024)Hong, Zhuge, Chen, Zheng, Cheng, Wang, Zhang, Wang, Yau, Lin, Zhou, Ran, Xiao, Wu, and Schmidhuber}]{hong2023metagpt}
Sirui Hong, Mingchen Zhuge, Jonathan Chen, Xiawu Zheng, Yuheng Cheng, Jinlin Wang, Ceyao Zhang, Zili Wang, Steven Ka~Shing Yau, Zijuan Lin, Liyang Zhou, Chenyu Ran, Lingfeng Xiao, Chenglin Wu, and J{\"{u}}rgen Schmidhuber. 2024.
\newblock \href {https://openreview.net/forum?id=VtmBAGCN7o} {Metagpt: Meta programming for {A} multi-agent collaborative framework}.
\newblock In \emph{The Twelfth International Conference on Learning Representations, {ICLR} 2024, Vienna, Austria, May 7-11, 2024}. OpenReview.net.

\bibitem[{Huang et~al.(2025{\natexlab{a}})Huang, Huang, Yu, Xie, Wu, Zhang, Mcauley, Jannach, and Yao}]{huang2025survey}
Chengkai Huang, Hongtao Huang, Tong Yu, Kaige Xie, Junda Wu, Shuai Zhang, Julian Mcauley, Dietmar Jannach, and Lina Yao. 2025{\natexlab{a}}.
\newblock A survey of foundation model-powered recommender systems: From feature-based, generative to agentic paradigms.
\newblock \emph{arXiv preprint arXiv:2504.16420}.

\bibitem[{Huang et~al.(2025{\natexlab{b}})Huang, Wu, Xie, Xia, Wang, Yu, Mitra, McAuley, and Yao}]{huang2025pluralistic}
Chengkai Huang, Junda Wu, Zhouhang Xie, Yu~Xia, Rui Wang, Tong Yu, Subrata Mitra, Julian McAuley, and Lina Yao. 2025{\natexlab{b}}.
\newblock Pluralistic off-policy evaluation and alignment.
\newblock \emph{arXiv preprint arXiv:2509.19333}.

\bibitem[{Huang et~al.(2025{\natexlab{c}})Huang, Xia, Wang, Xie, Yu, McAuley, and Yao}]{huang2025embedding}
Chengkai Huang, Yu~Xia, Rui Wang, Kaige Xie, Tong Yu, Julian McAuley, and Lina Yao. 2025{\natexlab{c}}.
\newblock Embedding-informed adaptive retrieval-augmented generation of large language models.
\newblock In \emph{Proceedings of the 31st International Conference on Computational Linguistics}, pages 1403--1412.

\bibitem[{Huang et~al.(2025{\natexlab{d}})Huang, Huang, Wu, Yu, McAuley, and Yao}]{huang2025listwise}
Hongtao Huang, Chengkai Huang, Junda Wu, Tong Yu, Julian McAuley, and Lina Yao. 2025{\natexlab{d}}.
\newblock \href {https://openreview.net/forum?id=x5KUOlYKQr} {Listwise preference diffusion optimization for user behavior trajectories prediction}.
\newblock In \emph{The Thirty-ninth Annual Conference on Neural Information Processing Systems}.

\bibitem[{Hurst et~al.(2024)Hurst, Lerer, Goucher, Perelman, Ramesh, Clark, Ostrow, Welihinda, Hayes, Radford, Madry, Baker{-}Whitcomb, Beutel, Borzunov, Carney, Chow, Kirillov, Nichol, Paino, Renzin, Passos, Kirillov, Christakis, Conneau, Kamali, Jabri, Moyer, Tam, Crookes, Tootoonchian, Kumar, Vallone, Karpathy, Braunstein, Cann, Codispoti, Galu, Kondrich, Tulloch, Mishchenko, Baek, Jiang, Pelisse, Woodford, Gosalia, Dhar, Pantuliano, Nayak, Oliver, Zoph, Ghorbani, Leimberger, Rossen, Sokolowsky, Wang, Zweig, Hoover, Samic, McGrew, Spero, Giertler, Cheng, Lightcap, Walkin, Quinn, Guarraci, Hsu, Kellogg, Eastman, Lugaresi, Wainwright, Bassin, Hudson, Chu, Nelson, Li, Shern, Conger, Barette, Voss, Ding, Lu, Zhang, Beaumont, Hallacy, Koch, Gibson, Kim, Choi, McLeavey, Hesse, Fischer, Winter, Czarnecki, Jarvis, Wei, Koumouzelis, and Sherburn}]{hurst2024gpt}
Aaron Hurst, Adam Lerer, Adam~P. Goucher, Adam Perelman, Aditya Ramesh, Aidan Clark, AJ~Ostrow, Akila Welihinda, Alan Hayes, Alec Radford, Aleksander Madry, Alex Baker{-}Whitcomb, Alex Beutel, Alex Borzunov, Alex Carney, Alex Chow, Alex Kirillov, Alex Nichol, Alex Paino, and 79 others. 2024.
\newblock \href {https://doi.org/10.48550/ARXIV.2410.21276} {Gpt-4o system card}.
\newblock \emph{CoRR}, abs/2410.21276.

\bibitem[{Imambi et~al.(2021)Imambi, Prakash, and Kanagachidambaresan}]{imambi2021pytorch}
Sagar Imambi, Kolla~Bhanu Prakash, and GR~Kanagachidambaresan. 2021.
\newblock Pytorch.
\newblock In \emph{Programming with TensorFlow: solution for edge computing applications}, pages 87--104. Springer.

\bibitem[{Jiao et~al.(2026)Jiao, Xiao, Wei, Qi, Huang, Sheng, and Yao}]{jiao2026prunerag}
Shuguang Jiao, Xinyu Xiao, Yunfan Wei, Shuhan Qi, Chengkai Huang, Quan Z~Michael Sheng, and Lina Yao. 2026.
\newblock Prunerag: Confidence-guided query decomposition trees for efficient retrieval-augmented generation.
\newblock \emph{arXiv preprint arXiv:2601.11024}.

\bibitem[{Kulkarni and Burhanpurwala(2024)}]{kulkarni2024survey}
Omkaresh Kulkarni and Adnan Burhanpurwala. 2024.
\newblock A survey of advancements in dbscan clustering algorithms for big data.
\newblock In \emph{2024 3rd International conference on Power Electronics and IoT Applications in Renewable Energy and its Control (PARC)}, pages 106--111. IEEE.

\bibitem[{Lee et~al.(2024)Lee, Chen, Furuta, Canny, and Fischer}]{lee2024human}
Kuang{-}Huei Lee, Xinyun Chen, Hiroki Furuta, John~F. Canny, and Ian Fischer. 2024.
\newblock \href {https://openreview.net/forum?id=OTmcsyEO5G} {A human-inspired reading agent with gist memory of very long contexts}.
\newblock In \emph{Forty-first International Conference on Machine Learning, {ICML} 2024, Vienna, Austria, July 21-27, 2024}. OpenReview.net.

\bibitem[{Lewis et~al.(2020)Lewis, Perez, Piktus, Petroni, Karpukhin, Goyal, K{\"{u}}ttler, Lewis, Yih, Rockt{\"{a}}schel, Riedel, and Kiela}]{lewis2020retrieval}
Patrick Lewis, Ethan Perez, Aleksandra Piktus, Fabio Petroni, Vladimir Karpukhin, Naman Goyal, Heinrich K{\"{u}}ttler, Mike Lewis, Wen{-}tau Yih, Tim Rockt{\"{a}}schel, Sebastian Riedel, and Douwe Kiela. 2020.
\newblock \href {https://proceedings.neurips.cc/paper/2020/hash/6b493230205f780e1bc26945df7481e5-Abstract.html} {Retrieval-augmented generation for knowledge-intensive {NLP} tasks}.
\newblock In \emph{Advances in Neural Information Processing Systems 33: Annual Conference on Neural Information Processing Systems 2020, NeurIPS 2020, December 6-12, 2020, virtual}.

\bibitem[{Li et~al.(2021)Li, Wang, Qi, Huang, Jiang, Liao, Guan, and Zhang}]{li2021self}
Yifan Li, Xuan Wang, Shuhan Qi, Chengkai Huang, Zoe~L Jiang, Qing Liao, Jian Guan, and Jiajia Zhang. 2021.
\newblock Self-supervised learning-based weight adaptive hashing for fast cross-modal retrieval.
\newblock \emph{Signal, Image and Video Processing}, 15(4):673--680.

\bibitem[{Liu et~al.(2023)Liu, Yang, Shen, Hu, Zhang, Gu, and Zhang}]{liu2023think}
Lei Liu, Xiaoyan Yang, Yue Shen, Binbin Hu, Zhiqiang Zhang, Jinjie Gu, and Guannan Zhang. 2023.
\newblock \href {https://doi.org/10.48550/ARXIV.2311.08719} {Think-in-memory: Recalling and post-thinking enable llms with long-term memory}.
\newblock \emph{CoRR}, abs/2311.08719.

\bibitem[{Liu et~al.(2021)Liu, Liu, and Huang}]{liu2021semi}
Zhiqiang Liu, Yanxia Liu, and Chengkai Huang. 2021.
\newblock Semi-online knowledge distillation.
\newblock \emph{arXiv preprint arXiv:2111.11747}.

\bibitem[{Maharana et~al.(2024)Maharana, Lee, Tulyakov, Bansal, Barbieri, and Fang}]{maharana2024evaluating}
Adyasha Maharana, Dong{-}Ho Lee, Sergey Tulyakov, Mohit Bansal, Francesco Barbieri, and Yuwei Fang. 2024.
\newblock \href {https://doi.org/10.18653/V1/2024.ACL-LONG.747} {Evaluating very long-term conversational memory of {LLM} agents}.
\newblock In \emph{Proceedings of the 62nd Annual Meeting of the Association for Computational Linguistics (Volume 1: Long Papers), {ACL} 2024, Bangkok, Thailand, August 11-16, 2024}, pages 13851--13870. Association for Computational Linguistics.

\bibitem[{Ouyang et~al.(2025)Ouyang, Yan, Hsu, Chen, Jiang, Wang, Han, Le, Daruki, Tang, Tirumalashetty, Lee, Rofouei, Lin, Han, Lee, and Pfister}]{ouyang2025reasoningbank}
Siru Ouyang, Jun Yan, I{-}Hung Hsu, Yanfei Chen, Ke~Jiang, Zifeng Wang, Rujun Han, Long~T. Le, Samira Daruki, Xiangru Tang, Vishy Tirumalashetty, George Lee, Mahsan Rofouei, Hangfei Lin, Jiawei Han, Chen{-}Yu Lee, and Tomas Pfister. 2025.
\newblock \href {https://doi.org/10.48550/ARXIV.2509.25140} {Reasoningbank: Scaling agent self-evolving with reasoning memory}.
\newblock \emph{CoRR}, abs/2509.25140.

\bibitem[{Packer et~al.(2023)Packer, Fang, Patil, Lin, Wooders, and Gonzalez}]{packer2023memgpt}
Charles Packer, Vivian Fang, Shishir~G. Patil, Kevin Lin, Sarah Wooders, and Joseph~E. Gonzalez. 2023.
\newblock \href {https://doi.org/10.48550/ARXIV.2310.08560} {Memgpt: Towards llms as operating systems}.
\newblock \emph{CoRR}, abs/2310.08560.

\bibitem[{Park et~al.(2023)Park, O'Brien, Cai, Morris, Liang, and Bernstein}]{park2023generative}
Joon~Sung Park, Joseph~C. O'Brien, Carrie~Jun Cai, Meredith~Ringel Morris, Percy Liang, and Michael~S. Bernstein. 2023.
\newblock \href {https://doi.org/10.1145/3586183.3606763} {Generative agents: Interactive simulacra of human behavior}.
\newblock In \emph{Proceedings of the 36th Annual {ACM} Symposium on User Interface Software and Technology, {UIST} 2023, San Francisco, CA, USA, 29 October 2023- 1 November 2023}, pages 2:1--2:22. {ACM}.

\bibitem[{Reimers and Gurevych(2019)}]{reimers2019sentence}
Nils Reimers and Iryna Gurevych. 2019.
\newblock \href {https://doi.org/10.18653/V1/D19-1410} {Sentence-bert: Sentence embeddings using siamese bert-networks}.
\newblock In \emph{Proceedings of the 2019 Conference on Empirical Methods in Natural Language Processing and the 9th International Joint Conference on Natural Language Processing, {EMNLP-IJCNLP} 2019, Hong Kong, China, November 3-7, 2019}, pages 3980--3990. Association for Computational Linguistics.

\bibitem[{Schubert et~al.(2017)Schubert, Sander, Ester, Kriegel, and Xu}]{schubert2017dbscan}
Erich Schubert, J{\"{o}}rg Sander, Martin Ester, Hans{-}Peter Kriegel, and Xiaowei Xu. 2017.
\newblock \href {https://doi.org/10.1145/3068335} {{DBSCAN} revisited, revisited: Why and how you should (still) use {DBSCAN}}.
\newblock \emph{{ACM} Trans. Database Syst.}, 42(3):19:1--19:21.

\bibitem[{Shinn et~al.(2023)Shinn, Cassano, Gopinath, Narasimhan, and Yao}]{shinn2023reflexion}
Noah Shinn, Federico Cassano, Ashwin Gopinath, Karthik Narasimhan, and Shunyu Yao. 2023.
\newblock \href {http://papers.nips.cc/paper\_files/paper/2023/hash/1b44b878bb782e6954cd888628510e90-Abstract-Conference.html} {Reflexion: language agents with verbal reinforcement learning}.
\newblock In \emph{Advances in Neural Information Processing Systems 36: Annual Conference on Neural Information Processing Systems 2023, NeurIPS 2023, New Orleans, LA, USA, December 10 - 16, 2023}.

\bibitem[{Sun et~al.(2024)Sun, Xu, Tang, Wang, Lin, Gong, Ni, Shum, and Guo}]{sun2023think}
Jiashuo Sun, Chengjin Xu, Lumingyuan Tang, Saizhuo Wang, Chen Lin, Yeyun Gong, Lionel~M. Ni, Heung{-}Yeung Shum, and Jian Guo. 2024.
\newblock \href {https://openreview.net/forum?id=nnVO1PvbTv} {Think-on-graph: Deep and responsible reasoning of large language model on knowledge graph}.
\newblock In \emph{The Twelfth International Conference on Learning Representations, {ICLR} 2024, Vienna, Austria, May 7-11, 2024}. OpenReview.net.

\bibitem[{Team(2024)}]{grattafiori2024llama}
Llama Team. 2024.
\newblock \href {https://doi.org/10.48550/ARXIV.2407.21783} {The llama 3 herd of models}.
\newblock \emph{CoRR}, abs/2407.21783.

\bibitem[{Wang et~al.(2023)Wang, Liang, Yang, Huang, Wu, Wu, Lu, Ma, and Li}]{wang2023scm}
Bing Wang, Xinnian Liang, Jian Yang, Hui Huang, Shuangzhi Wu, Peihao Wu, Lu~Lu, Zejun Ma, and Zhoujun Li. 2023.
\newblock Scm: Enhancing large language model with self-controlled memory framework.
\newblock \emph{arXiv e-prints}, pages arXiv--2304.

\bibitem[{Wu et~al.(2025)Wu, Liu, Choi, and Shu}]{wu2025improving}
Shanglin Wu, Lihui Liu, Jinho~D. Choi, and Kai Shu. 2025.
\newblock \href {https://doi.org/10.48550/ARXIV.2509.03540} {Improving factuality in llms via inference-time knowledge graph construction}.
\newblock \emph{CoRR}, abs/2509.03540.

\bibitem[{Xu et~al.(2025)Xu, Liang, Mei, Gao, Tan, and Zhang}]{xu2025mem}
Wujiang Xu, Zujie Liang, Kai Mei, Hang Gao, Juntao Tan, and Yongfeng Zhang. 2025.
\newblock \href {https://doi.org/10.48550/ARXIV.2502.12110} {{A-MEM:} agentic memory for {LLM} agents}.
\newblock \emph{CoRR}, abs/2502.12110.

\bibitem[{Yang et~al.(2025)Yang, Yu, Li, Liu, Huang, Huang, Jiang, Tu, Zhang, Zhou, Lin, Dang, Yang, Yu, Li, Sun, Zhu, Men, He, Xu, Yin, Yu, Qiu, Ren, Yang, Li, Xu, and Zhang}]{yang2025qwen2}
An~Yang, Bowen Yu, Chengyuan Li, Dayiheng Liu, Fei Huang, Haoyan Huang, Jiandong Jiang, Jianhong Tu, Jianwei Zhang, Jingren Zhou, Junyang Lin, Kai Dang, Kexin Yang, Le~Yu, Mei Li, Minmin Sun, Qin Zhu, Rui Men, Tao He, and 9 others. 2025.
\newblock \href {https://doi.org/10.48550/ARXIV.2501.15383} {Qwen2.5-1m technical report}.
\newblock \emph{CoRR}, abs/2501.15383.

\bibitem[{Yang et~al.(2018)Yang, Qi, Zhang, Bengio, Cohen, Salakhutdinov, and Manning}]{yang2018hotpotqa}
Zhilin Yang, Peng Qi, Saizheng Zhang, Yoshua Bengio, William~W. Cohen, Ruslan Salakhutdinov, and Christopher~D. Manning. 2018.
\newblock \href {https://doi.org/10.18653/V1/D18-1259} {Hotpotqa: {A} dataset for diverse, explainable multi-hop question answering}.
\newblock In \emph{Proceedings of the 2018 Conference on Empirical Methods in Natural Language Processing, Brussels, Belgium, October 31 - November 4, 2018}, pages 2369--2380. Association for Computational Linguistics.

\bibitem[{Yao et~al.(2023)Yao, Zhao, Yu, Du, Shafran, Narasimhan, and Cao}]{yao2022react}
Shunyu Yao, Jeffrey Zhao, Dian Yu, Nan Du, Izhak Shafran, Karthik~R. Narasimhan, and Yuan Cao. 2023.
\newblock \href {https://openreview.net/forum?id=WE\_vluYUL-X} {React: Synergizing reasoning and acting in language models}.
\newblock In \emph{The Eleventh International Conference on Learning Representations, {ICLR} 2023, Kigali, Rwanda, May 1-5, 2023}. OpenReview.net.

\bibitem[{Yasunaga et~al.(2021)Yasunaga, Ren, Bosselut, Liang, and Leskovec}]{yasunaga2021qa}
Michihiro Yasunaga, Hongyu Ren, Antoine Bosselut, Percy Liang, and Jure Leskovec. 2021.
\newblock \href {https://doi.org/10.18653/V1/2021.NAACL-MAIN.45} {{QA-GNN:} reasoning with language models and knowledge graphs for question answering}.
\newblock In \emph{Proceedings of the 2021 Conference of the North American Chapter of the Association for Computational Linguistics: Human Language Technologies, {NAACL-HLT} 2021, Online, June 6-11, 2021}, pages 535--546. Association for Computational Linguistics.

\bibitem[{Zhang et~al.(2025)Zhang, Fu, Wan, Yu, Wang, and Yan}]{zhang2025g}
Guibin Zhang, Muxin Fu, Guancheng Wan, Miao Yu, Kun Wang, and Shuicheng Yan. 2025.
\newblock \href {https://doi.org/10.48550/ARXIV.2506.07398} {G-memory: Tracing hierarchical memory for multi-agent systems}.
\newblock \emph{CoRR}, abs/2506.07398.

\bibitem[{Zhong et~al.(2024{\natexlab{a}})Zhong, Wu, Li, Peng, and Wu}]{zhong2023comprehensive}
Lingfeng Zhong, Jia Wu, Qian Li, Hao Peng, and Xindong Wu. 2024{\natexlab{a}}.
\newblock \href {https://doi.org/10.1145/3618295} {A comprehensive survey on automatic knowledge graph construction}.
\newblock \emph{{ACM} Comput. Surv.}, 56(4):94:1--94:62.

\bibitem[{Zhong et~al.(2024{\natexlab{b}})Zhong, Guo, Gao, Ye, and Wang}]{zhong2024memorybank}
Wanjun Zhong, Lianghong Guo, Qiqi Gao, He~Ye, and Yanlin Wang. 2024{\natexlab{b}}.
\newblock \href {https://doi.org/10.1609/AAAI.V38I17.29946} {Memorybank: Enhancing large language models with long-term memory}.
\newblock In \emph{Thirty-Eighth {AAAI} Conference on Artificial Intelligence, {AAAI} 2024, Thirty-Sixth Conference on Innovative Applications of Artificial Intelligence, {IAAI} 2024, Fourteenth Symposium on Educational Advances in Artificial Intelligence, {EAAI} 2014, February 20-27, 2024, Vancouver, Canada}, pages 19724--19731. {AAAI} Press.

\end{thebibliography}

\appendix
\newpage
\onecolumn

\section*{Appendix Outline}
\label{appendix:outline}
{
\hypersetup{hidelinks}
}

\noindent
A.\,\hyperref[app:env]{Environment Details}\dotfill\pageref{app:env}\\
B.\,\hyperref[app:impl]{Implementation Details}\dotfill\pageref{app:impl}\\
\hspace*{3em}Dataset and Evaluation Scope.\dotfill\pageref{app:impl:dataset}\\
\hspace*{3em}Backbone Models and Inference.\dotfill\pageref{app:impl:backbone}\\
\hspace*{3em}Offline Memory Construction.\dotfill\pageref{app:impl:offl}\\
\hspace*{3em}Graph Memory Construction.\dotfill\pageref{app:impl:gm}\\
\hspace*{3em}Experience Memory Induction.\dotfill\pageref{app:impl:em}\\
\hspace*{3em}Retrieval and Context Assembly.\dotfill\pageref{app:impl:retrie}\\
C.\,\hyperref[app:prompt]{Prompt Templates}\dotfill\pageref{app:prompt}\\
\hspace*{3em}Entity Extraction Prompt\dotfill\pageref{app:prompt:ent}\\
\hspace*{3em}Relation Extraction Prompt\dotfill\pageref{app:prompt:rel}\\
\hspace*{3em}Temporal Normalization Prompt\dotfill\pageref{app:prompt:temp}\\
\hspace*{3em}Knowledge Graph Verification Prompt\dotfill\pageref{app:prompt:kg}\\
\hspace*{3em}Experience Extraction Prompt\dotfill\pageref{app:prompt:exp}\\
D.\,\hyperref[app:experiment]{Experiment}\dotfill\pageref{app:experiment}\\
\hspace*{1.5em}D.1\,\hyperref[app:experiment:detail]{Detailed Baselines Introduction}\dotfill\pageref{app:experiment:detail}\\
\hspace*{1.5em}D.2\,\hyperref[app:experiment:eval]{Evaluation Metric}\dotfill\pageref{app:experiment:eval}\\
\hspace*{1.5em}D.3\,\hyperref[app:experiment:deepseek]{DeepSeek-V3.2 Backbone}\dotfill\pageref{app:experiment:deepseek}\\
\hspace*{1.5em}D.4\,\hyperref[app:experiment:comp]{Comparison Results}\dotfill\pageref{app:experiment:comp}\\
\hspace*{1.5em}D.5\,\hyperref[app:experiment:case]{Case Study}\dotfill\pageref{app:experiment:case}\\
\hspace*{1.5em}D.6\,\hyperref[app:experiment:scale]{Model Scaling Analysis}\dotfill\pageref{app:experiment:scale}\\
\hspace*{1.5em}D.7\,\hyperref[app:experiment:human]{Human Evaluation Details}\dotfill\pageref{app:experiment:human}\\
E.\,\hyperref[app:algorithm]{Algorithm}\dotfill\pageref{app:algorithm}\\
\hspace*{1.5em}E.1\,\hyperref[app:algorithm-cons]{Memory Construction}\dotfill\pageref{app:algorithm-cons}\\
\hspace*{3em}KG Writing and Maintenance\dotfill\pageref{alg:kg-build}\\
\hspace*{3em}Experience Induction and Online Update\dotfill\pageref{alg:exp-update}\\
\hspace*{1.5em}E.2\,\hyperref[app:algorithm-retrie]{Retrieval and Reasoning}\dotfill\pageref{app:algorithm-retrie}\\
\hspace*{3em}Inference Retrieval and Context Assembly\dotfill\pageref{alg:retrieve}\\


\twocolumn
\newpage
\section{Environment Details}\label{app:env}

All experiments were conducted on a Linux server running
\texttt{Linux 6.8.0-90-generic} with \texttt{glibc 2.39}.
The software stack was based on \texttt{Python 3.10.19} and
\texttt{PyTorch 2.9.0} \cite{imambi2021pytorch} compiled with \texttt{CUDA 12.8} (\texttt{torch 2.9.0+cu128}),
with \texttt{cuDNN 9.10.2} enabled.
All experiments were executed with GPU acceleration.

The hardware configuration consisted of four
\texttt{NVIDIA A100-SXM4 GPUs} with \texttt{80GB} HBM2e memory each.
The installed NVIDIA driver version was \texttt{575.57.08},
and the CUDA runtime reported by the driver was \texttt{CUDA 12.9}.
Unless otherwise specified, all memory construction and inference
pipelines were executed on this hardware configuration.

\section{Implementation Details}\label{app:impl}

\paragraph{Dataset and Evaluation Scope.}\label{app:impl:dataset}
We evaluate MemWeaver on the LoCoMo dataset released by Snap Research,\footnote{\url{https://github.com/snap-research/locomo}}
which is designed for long-term conversational question answering over multi-session dialogue histories. LoCoMo categorizes questions into five types:
(1) \textit{Single-Hop} questions answerable from a single session;
(2) \textit{Multi-Hop} questions requiring information synthesis across sessions;
(3) \textit{Temporal} questions that test time-aware reasoning;
(4) \textit{Open-Domain} questions that require integrating conversational context
with general knowledge; and
(5) \textit{Adversarial} questions that are unanswerable from the dialogue history.
In total, the dataset contains 7,512 question--answer pairs across these categories.
Since our primary goal is to evaluate long-term memory and evidence-supported reasoning over extended dialogue trajectories, we mainly focus on the four answerable categories (Single-Hop, Multi-Hop, Temporal, and Open-Domain) in our main experiments.

The dataset does not provide an official train/dev/test split, and following prior work, we evaluate directly on the full set of annotated question--answer pairs without further partitioning.
In our main experiments, we focus on the four answerable question categories:
\emph{Single-Hop}, \emph{Multi-Hop}, \emph{Temporal}, and \emph{Open-Domain}.
Adversarial questions are excluded in the main experiment part, as they primarily evaluate abstention behavior rather than evidence-supported reasoning. However, we still report the experiments on \emph{Adversarial} dataset in Appendix \ref{app:adv}. 
We do not fix a random seed, since the system does not involve stochastic training and all memory construction is performed deterministically given the underlying LLM outputs.

\paragraph{Backbone Models and Inference.}\label{app:impl:backbone}
We evaluate four backbone language models:
GPT-4o-mini \cite{hurst2024gpt}, Llama3.2-3B \cite{grattafiori2024llama}, Llama3.2-1B \cite{grattafiori2024llama}, and Qwen2.5-1.5B \cite{yang2025qwen2}.
GPT-4o-mini is accessed via a commercial API, while the remaining models are served locally using \texttt{Ollama}.
All methods, including baselines and MemWeaver, share identical system prompts and output formats to ensure fair comparison.
Category-specific answer prompts are used at inference time to accommodate differences in question styles (e.g., temporal or multi-hop queries), while keeping the overall prompting strategy consistent across methods.
Decoding configurations follow standard practice for each backend and are kept consistent within each model.

\paragraph{Offline Memory Construction.}\label{app:impl:offl}
All memory components in MemWeaver are constructed offline prior to inference.
We employ DeepSeek-V3.2 (API-based) \cite{deepseek2025deepseek} exclusively for memory construction, including entity extraction, relation extraction, experience induction, and session-level verification.
This model is \emph{not} used during inference.
Once the full memory is built, it remains fixed throughout evaluation, and all backbone models query the same memory state.

Textual embeddings for dialogue units, passages, experience items, and knowledge graph triples are computed using the \texttt{all-minilm-l6-v2} \cite{reimers2019sentence} sentence encoder. The same embedding model is used consistently across clustering, routing, and retrieval.

\paragraph{Graph Memory Construction.}\label{app:impl:gm}
Graph Memory is implemented as a directed knowledge graph that consolidates relational facts across dialogue sessions.
Entity extraction, relation extraction, and session-level review are performed using fixed prompt templates.
Implicit temporal expressions in dialogue are normalized into absolute time representations during memory writing.
Structured retrieval operates over semantic relation triples only, excluding structural edges.
During inference, graph retrieval expands candidate triples using a single-hop neighborhood expansion from seed relations.

\paragraph{Experience Memory Induction.}\label{app:impl:em}
Experience Memory is constructed by clustering dialogue units using DBSCAN with cosine distance.
We set the clustering parameters to $\texttt{eps}=0.3$ and $\texttt{min\_samples}=2$.
Each candidate cluster is screened for semantic coherence before inducing reusable experience items.
For online updates, newly arriving dialogue units are routed to existing clusters based on cosine similarity, with thresholds $\texttt{sim\_high}=0.8$ and $\texttt{sim\_low}=0.5$.
Each cluster maintains an update buffer, and experience re-induction is triggered when the buffer size reaches 4, amortizing the cost of LLM-based updates in long-running settings.

\paragraph{Retrieval and Context Assembly.}\label{app:impl:retrie}
MemWeaver employs a dual-channel retrieval strategy.
Structured retrieval over Graph Memory provides relational facts, while textual retrieval gathers supporting passages and experience items.
Unless otherwise specified, the retrieval budgets are fixed to $k_r = k_p = k_e = 6$. For graph triples, passages, and experience items, respectively, across all question categories.
The retrieved structured and textual contexts are assembled into a compact inference input, which is then provided to the backbone language model for answer generation.

\clearpage
\onecolumn
\section{Prompt Templates}\label{app:prompt}
\begin{figure}[H]
\centering
\begin{tcolorbox}[
  enhanced,
  colback=blue!3,
  colframe=blue!30!black,
  arc=6pt,
  boxrule=0.8pt,
  width=\linewidth,
  title=\textbf{Entity Extraction Prompt},
  coltitle=black,
  colbacktitle=blue!25,
  boxed title style={
    colframe=blue!30!black,
    colback=blue!25,
    arc=6pt
  },
  attach boxed title to top left={
    xshift=6pt,
    yshift*=-\dimexpr\tcboxedtitleheight/2\relax
  },
  fontupper=\small,
  top=6pt,bottom=6pt,left=8pt,right=8pt
]

\textbf{You are an entity extraction assistant.}\\
\textbf{Your task:} Extract entities from a dialogue snippet.

\vspace{0.5em}
\textbf{Guidelines:}
\begin{itemize}
  \item Extract concise entity names that appear in the text (people, locations, organizations, events, objects, etc.).
  \item Do not invent entities.
  \item Prefer a canonical form of the entity name.
  \item If an entity is expressed as a combined phrase, keep the full phrase intact and do not split it into smaller parts.
  \item Return a de-duplicated list.
\end{itemize}

\vspace{0.2em}
\textbf{DO NOT EXTRACT:}
\begin{enumerate}
  \item Unclear entities such as \texttt{"he"}, \texttt{"that"}, \texttt{"there"}.
  \item Time/Date expressions (relative or absolute) such as \texttt{"yesterday"}, \texttt{"last week"}, \texttt{"next month"}.
\end{enumerate}

\vspace{0.5em}
\textbf{Dialogue:}\\
\texttt{\{dialogue\_text\}}

\vspace{0.5em}
\textbf{Output (STRICT JSON):}
\begin{lstlisting}
{
  "entities": ["entity1", "entity2", ...]
}
\end{lstlisting}

\end{tcolorbox}
\caption{Prompt for entity extraction from a dialogue snippet.}
\label{app:prompt:ent}
\end{figure}

\begin{figure*}[!htbp]
\centering
\begin{tcolorbox}[
  enhanced,
  colback=blue!3,
  colframe=blue!30!black,
  arc=6pt,
  boxrule=0.8pt,
  width=\linewidth,
  title=\textbf{Relation Extraction Prompt},
  coltitle=black,
  colbacktitle=blue!25,
  boxed title style={
    colframe=blue!30!black,
    colback=blue!25,
    arc=6pt
  },
  attach boxed title to top left={
    xshift=6pt,
    yshift*=-\dimexpr\tcboxedtitleheight/2\relax
  },
  fontupper=\small,
  top=6pt,bottom=6pt,left=8pt,right=8pt
]

\textbf{You are a relation extraction assistant.}\\
\textbf{Your task:} Extract meaningful relations between entities as triples from a dialogue snippet.

\vspace{0.5em}
\textbf{Guidelines:}
\begin{itemize}
  \item Extract relations \textbf{ONLY} between entities in the provided list.
  \item Each relation must be directly supported by the text.
  \item \texttt{relation\_type} should be a short predicate phrase (\emph{lowercase preferred}).
  \item If the relation happens under certain conditions, you can optionally include a \texttt{"condition"} field to describe it briefly.
\end{itemize}

\vspace{0.2em}
\textbf{DO NOT EXTRACT:}
\begin{itemize}
  \item time/location
  \item unmeaningful or vague relations, e.g., \texttt{"is related to"}, \texttt{"has something to do with"}, \texttt{"is associated with"}, etc.
\end{itemize}

\vspace{0.5em}
\textbf{Dialogue:}\\
\texttt{\{dialogue\_text\}}

\vspace{0.5em}
\textbf{Detected entities:}\\
\texttt{\{entity\_list\_text\}}

\vspace{0.5em}
\textbf{Output (STRICT JSON):}
\begin{lstlisting}
{
  "relations": [
    {
      "source": "entity_name1",
      "target": "entity_name2",
      "relation_type": "short relation phrase",
      "condition": "if mentioned"
    }
  ]
}
\end{lstlisting}

\end{tcolorbox}

\caption{Prompt for relation extraction between detected entities in a dialogue snippet.}
\label{app:prompt:rel}
\end{figure*}
\FloatBarrier
\clearpage

\begin{figure}[!htbp]
\centering
\begin{tcolorbox}[
  enhanced,
  colback=blue!3,
  colframe=blue!30!black,
  arc=6pt,
  boxrule=0.8pt,
  width=\linewidth,
  title=\textbf{Temporal Normalization Prompt},
  coltitle=black,
  colbacktitle=blue!25,
  boxed title style={
    colframe=blue!30!black,
    colback=blue!25,
    arc=6pt
  },
  attach boxed title to top left={
    xshift=6pt,
    yshift*=-\dimexpr\tcboxedtitleheight/2\relax
  },
  fontupper=\small,
  top=6pt,bottom=6pt,left=8pt,right=8pt
]

\textbf{You are a time expression extractor.}\\
\textbf{Your task:} Extract an \textbf{ABSOLUTE} time expression indicating when the event described by the given relation occurred or is scheduled to occur; the absolute time may be inferred from relative time expressions in the dialogue.

\vspace{0.5em}
\textbf{Guidelines:}
\begin{itemize}
  \item Identify the time \textbf{MOST relevant} to the given relation/event.
  \item The dialogue may contain:
  \begin{itemize}
    \item Absolute time expressions (e.g., \texttt{"20 May 2022"}, \texttt{"May 2022"}, \texttt{"2022"})
    \item Relative time expressions (e.g., \texttt{"yesterday"}, \texttt{"last week"}, \texttt{"this weekend"})
  \end{itemize}
  \item If a relative time expression is present, you \textbf{MAY} use it as a clue and resolve it into an \textbf{ABSOLUTE} time.
  \item The final output \textbf{MUST} be an \textbf{ABSOLUTE, HUMAN-READABLE} time expression.
\end{itemize}

\vspace{0.2em}
\textbf{What counts as HUMAN-READABLE ABSOLUTE time:}
\begin{itemize}
  \item A specific calendar date (e.g., \texttt{"20 May, 2022"})
  \item A specific month and year (e.g., \texttt{"May, 2022"})
  \item A specific year (e.g., \texttt{"2022"})
\end{itemize}

\vspace{0.2em}
\textbf{DO NOT OUTPUT:}
\begin{itemize}
  \item Relative time expressions (e.g., \texttt{"yesterday"}, \texttt{"last week"}, \texttt{"tomorrow"})
  \item Vague time references without a calendar anchor (e.g., \texttt{"recently"}, \texttt{"soon"}, \texttt{"later"})
\end{itemize}

\vspace{0.2em}
If there is \textbf{NO} clear usable time information for the target relation, return an empty string \texttt{""}.

\vspace{0.5em}
\textbf{Output format MUST be one of:}
\begin{itemize}
  \item Day-level: \hspace{1.6em}\texttt{"20 May, 2022"}
  \item Month-level: \texttt{"May, 2022"}
  \item Year-level: \hspace{1.2em}\texttt{"2022"}
\end{itemize}

\vspace{0.5em}
\textbf{Given:}
\begin{itemize}
  \item Dialogue: \texttt{\{dialogue\_text\}}
  \item Relation: \texttt{\{relation\_desc\}}
\end{itemize}

\vspace{0.5em}
\textbf{Output (STRICT JSON):}
\begin{lstlisting}
{
  "absolute_time": ""
}
\end{lstlisting}

\end{tcolorbox}
\caption{Prompt for temporal normalization of dialogue-based relations into absolute time.}
\label{app:prompt:temp}
\end{figure}

\FloatBarrier
\clearpage

\begin{figure*}[!htbp]
\centering
\begin{tcolorbox}[
  enhanced,
  colback=blue!3,
  colframe=blue!30!black,
  arc=6pt,
  boxrule=0.8pt,
  width=\linewidth,
  title=\textbf{Knowledge Graph Verification Prompt},
  coltitle=black,
  colbacktitle=blue!25,
  boxed title style={
    colframe=blue!30!black,
    colback=blue!25,
    arc=6pt
  },
  attach boxed title to top left={
    xshift=6pt,
    yshift*=-\dimexpr\tcboxedtitleheight/2\relax
  },
  fontupper=\small,
  top=6pt,bottom=6pt,left=8pt,right=8pt
]

\textbf{You are a knowledge graph review assistant.}

\vspace{0.2em}
\textbf{Your tasks:}\\
Review a knowledge graph extracted from a dialogue session.

\vspace{0.5em}
\textbf{Options:}
\begin{enumerate}
  \item \textbf{ADD:} Find important relations that are clearly expressed in the dialogue but are \textbf{MISSING} from the current relation list.
  \item \textbf{UPDATE:} For some \textbf{EXISTING} relations, refine \texttt{relation\_type}, \texttt{time}/\texttt{condition} metadata.
  \item \textbf{DENY:} If a relation in the current list is clearly \textbf{NOT supported} or \textbf{contradicted} by the dialogue, mark it as denied (to be removed).
\end{enumerate}

\vspace{0.5em}
\textbf{You are given:}\\
\textbf{The full dialogue text for this session:}\\
The dialogue happened at: \texttt{\{dialogue\_timestamp\}}\\
\texttt{\{full\_dialogue\_text\}}

\vspace{0.5em}
\textbf{Existing entities in the KG for this session:}\\
\texttt{\{entities\_text\}}

\vspace{0.5em}
\textbf{Existing relations (triples) in the KG for this session:}\\
\texttt{\{relations\_text\}}

\vspace{0.5em}
\textbf{Output (STRICT JSON):}
\begin{lstlisting}
{
  "add": [
    {"source": "A", "relation_type": "predicate", "target": "B", "time": "if mentioned", "condition": "if mentioned"}
  ],
  "update": [
    {"relation_id": "rid", "relation_type": "new predicate", "time": "if mentioned", "condition": "if mentioned"}
  ],
  "deny": [
    {"relation_id": "rid"}
  ]
}
\end{lstlisting}

\end{tcolorbox}
\caption{Prompt for verifying and refining a dialogue-derived knowledge graph.}
\label{app:prompt:kg}
\end{figure*}

\FloatBarrier
\clearpage

\begin{figure*}[!htbp]
\centering
\begin{tcolorbox}[
  enhanced,
  colback=blue!3,
  colframe=blue!30!black,
  arc=6pt,
  boxrule=0.8pt,
  width=\linewidth,
  title=\textbf{Experience Extraction Prompt},
  coltitle=black,
  colbacktitle=blue!25,
  boxed title style={
    colframe=blue!30!black,
    colback=blue!25,
    arc=6pt
  },
  attach boxed title to top left={
    xshift=6pt,
    yshift*=-\dimexpr\tcboxedtitleheight/2\relax
  },
  fontupper=\small,
  top=6pt,bottom=6pt,left=8pt,right=8pt
]

\textbf{You extract reusable experiences from short dialogues.}\\
You will see several Q\&A items from the same semantic cluster. Each item has an index like \texttt{[0]}, \texttt{[1]}, etc.

\vspace{0.5em}
\textbf{Dialogue samples:}\\
\texttt{\{qa\_context\}}

\vspace{0.5em}
\textbf{Your task:}
\begin{itemize}
  \item Extract a \textbf{SMALL SET} of reusable experiences that can help in similar future cases.
  \item Each experience must be directly supported by the dialogue (do not invent facts).
  \item Do \textbf{NOT} output vague life advice or generic statements.
  \item Prefer \texttt{"fact"} or \texttt{"preference"} when the dialogue states something directly.
  \item Only use \texttt{"strategy"} when the experience clearly generalizes beyond this specific person.
  \item Avoid generic strategies like \texttt{"communicate more"}, \texttt{"be kind"}, or \texttt{"support is important"}.
  \item Only mark Q\&A indices where the experience is explicitly expressed.
  \item It is better to output fewer, high-quality experiences than many weak or generic ones.
  \item \texttt{content} $\le$ 120 characters.
\end{itemize}

\vspace{0.2em}
\textbf{Allowed types:}
\begin{itemize}
  \item \texttt{fact}: a stable fact likely to remain true
  \item \texttt{strategy}: a general reusable approach
  \item \texttt{preference}: a stable interest or habit
\end{itemize}

\vspace{0.5em}
\textbf{Few-shot example:}
\begin{lstlisting}
[0] Speaker=Alex
    Q: I started weekly therapy recently.
    A:
[1] Speaker=Ben
    Q: Has it helped?
    A:
[2] Speaker=Alex
    Q: Yes, talking regularly helps me feel less overwhelmed.
    A:
\end{lstlisting}

\vspace{0.2em}
\textbf{Example output:}
\begin{lstlisting}
{
  "experiences": [
    {
      "type": "fact",
      "content": "Alex attends weekly therapy and feels less overwhelmed.",
      "source_qa_indices": [0, 2]
    }
  ]
}
\end{lstlisting}

\vspace{0.5em}
Now output \textbf{STRICT JSON} for the current dialogue only:

\vspace{0.25em}
\textbf{Output (STRICT JSON):}
\begin{lstlisting}
{
  "experiences": [
    {
      "type": "fact | strategy | preference",
      "content": "short experience (<=120 chars)",
      "source_qa_indices": [0, 1]
    }
  ]
}
\end{lstlisting}

\end{tcolorbox}
\caption{Prompt for extracting reusable experiences from clustered Q\&A dialogue samples.}
\label{app:prompt:exp}
\end{figure*}

\FloatBarrier

\twocolumn
\section{Experiment}\label{app:experiment}

\subsection{Detailed Baselines Introduction}\label{app:experiment:detail}

LoCoMo~\cite{maharana2024evaluating} takes a direct approach by leveraging foundation models without memory mechanisms
for question answering tasks. For each query, it incorporates the complete preceding conversation
and questions into the prompt, evaluating the model’s reasoning capabilities.

ReadAgent~\cite{lee2024human} tackles long-context document processing through a sophisticated three-step methodology:
it begins with episode pagination to segment content into manageable chunks, followed by
memory gisting to distill each page into concise memory representations, and concludes with interactive look-up
to retrieve pertinent information as needed.

MemoryBank~\cite{zhong2024memorybank} introduces an innovative memory management system that maintains and efficiently retrieves historical interactions.
The system features a dynamic memory updating mechanism based on the Ebbinghaus Forgetting Curve theory,
which intelligently adjusts memory strength according to time and significance.
Additionally, it incorporates a user portrait building system that progressively refines its understanding of user personality
through continuous interaction analysis.

A-Mem~\cite{xu2025mem} proposes an agentic memory framework that constructs and maintains atomic memory units for long-horizon interactions.
It organizes memories into interconnected notes that can be incrementally updated across sessions,
enabling the agent to retrieve and reuse relevant memory entries when answering queries.

\subsection{Evaluation Metric}\label{app:experiment:eval}

The F1 score represents the harmonic mean of precision and recall, offering a balanced metric that
combines both measures into a single value. This metric is particularly valuable when balancing
between complete and accurate responses:
\begin{equation}
\mathrm{F1} = 2 \cdot \frac{\mathrm{precision} \cdot \mathrm{recall}}{\mathrm{precision} + \mathrm{recall}},
\end{equation}
where
\begin{equation}
\mathrm{precision} = \frac{\text{true positives}}{\text{true positives} + \text{false positives}},
\end{equation}
and
\begin{equation}
\mathrm{recall} = \frac{\text{true positives}}{\text{true positives} + \text{false negatives}}.
\end{equation}
In question-answering systems, the F1 score plays a crucial role in evaluating exact matches between
predicted and reference answers. This is especially important for span-based QA tasks, where systems
must identify precise text segments while maintaining comprehensive coverage of the answer.

\begin{table*}[t]
\centering
\scriptsize
\setlength{\tabcolsep}{3.2pt}
\renewcommand{\arraystretch}{1.05}

\newcommand{\g}[1]{\cellcolor{gray!20}{#1}}
\resizebox{0.92\textwidth}{!}{%
\begin{tabular}{l|c c c c c c c c c c c c}
\toprule
\multirow{2}{*}{\textbf{Method}}
& \multicolumn{3}{c}{\textbf{Multi-Hop}} &
\multicolumn{3}{c}{\textbf{Temporal}} &
\multicolumn{3}{c}{\textbf{Open-Domain}} &
\multicolumn{3}{c}{\textbf{Single-Hop}}\\
& \textbf{F1} & \textbf{BLEU} & \textbf{RGE-2} & \textbf{F1} & \textbf{BLEU} & \textbf{RGE-2} & \textbf{F1} & \textbf{BLEU} &
\textbf{RGE-2} & \textbf{F1} & \textbf{BLEU} & \textbf{RGE-2} \\
\midrule

MemoryBank
& 5.03 & 4.28 & 1.59
& 2.43 & 1.60 & 0.46
& 5.54 & 3.12 & 1.04
& 6.84 & 3.62 & 2.91\\
A-Mem
& 31.13 & 19.54 & 11.39 
& 41.36 & 31.90 & 15.67
& 12.31 & 10.66 & 3.03
& 41.58 & 36.69 & 26.69\\
\g{\textbf{MemWeaver}}
& \g{\textbf{31.35}} & \g{\textbf{20.29}} & \g{\textbf{11.83}}
& \g{\textbf{55.52}} & \g{\textbf{45.36}} & \g{\textbf{26.97}}
& \g{\textbf{21.13}} & \g{\textbf{16.34}} & \g{\textbf{5.90}}
& \g{\textbf{45.19}} & \g{\textbf{39.47}} & \g{\textbf{28.96}}\\
\bottomrule
\end{tabular}%
}
\caption{Experimental results on the LoCoMo dataset (DeepSeek-V3.2 backbone). Results are reported in F1 and BLEU-1 (\%). Best results in each row are in bold, and MemWeaver is highlighted in gray.}
\label{tab:deepseek}
\end{table*}

BLEU-1 evaluates the precision of unigram matches between system outputs
and reference texts:
\begin{equation}
\mathrm{BLEU\text{-}1} = \mathrm{BP} \cdot \exp\!\left( \sum_{n=1}^{1} w_n \log p_n \right),
\end{equation}
where the brevity penalty $\mathrm{BP}$ is defined as
\begin{equation}
\mathrm{BP} =
\begin{cases}
1, & c > r, \\
e^{1-r/c}, & c \le r,
\end{cases}
\end{equation}
and
\begin{equation}
p_n =
\frac{\sum_i \sum_k \min(h_{ik}, m_{ik})}{\sum_i \sum_k h_{ik}}.
\end{equation}
Here, $c$ is the candidate length, $r$ is the reference length, $h_{ik}$ is the count of the $n$-gram
$i$ in candidate $k$, and $m_{ik}$ is the maximum count of that $n$-gram in any reference.
In QA tasks, BLEU-1 evaluates lexical precision and is particularly useful for generative QA systems
where exact matching may be overly strict.

ROUGE-L measures the longest common subsequence (LCS) between the generated
and reference texts:
\begin{equation}
\mathrm{ROUGE\text{-}L} =
\frac{(1+\beta^2) R_l P_l}{R_l + \beta^2 P_l},
\end{equation}
where
\begin{equation}
R_l = \frac{\mathrm{LCS}(X,Y)}{|X|}, \qquad
P_l = \frac{\mathrm{LCS}(X,Y)}{|Y|}.
\end{equation}
ROUGE-2 computes bigram overlap between the generated and reference texts:
\begin{equation}
\mathrm{ROUGE\text{-}2} =
\frac{\sum_{\text{bigram} \in \text{ref}} \min(\mathrm{Count}_{\text{ref}}, \mathrm{Count}_{\text{cand}})}
{\sum_{\text{bigram} \in \text{ref}} \mathrm{Count}_{\text{ref}}}.
\end{equation}
ROUGE-L focuses on sequence-level matching, while ROUGE-2 emphasizes local word order.
Both metrics are useful for evaluating the fluency and coherence of generated answers.

METEOR computes a score based on aligned unigrams between candidate and
reference texts, accounting for synonyms and paraphrases:
\begin{equation}
\mathrm{METEOR} = F_{\text{mean}} \cdot (1 - \mathrm{Penalty}),
\end{equation}
where
\begin{equation}
F_{\text{mean}} = \frac{10PR}{R + 9P},
\end{equation}
and
\begin{equation}
\mathrm{Penalty} = 0.5 \cdot \left( \frac{ch}{m} \right)^3.
\end{equation}
Here, $P$ and $R$ denote precision and recall, $ch$ is the number of chunks, and $m$ is the number of
matched unigrams. METEOR captures semantic similarity beyond exact matching and is well suited for
evaluating paraphrased answers.

SBERT Similarity measures semantic similarity between two texts using
sentence embeddings:
\begin{equation}
\mathrm{SBERT}(x,y) = \cos(\mathbf{e}_x, \mathbf{e}_y)
= \frac{\mathbf{e}_x \cdot \mathbf{e}_y}{\|\mathbf{e}_x\| \|\mathbf{e}_y\|},
\end{equation}
where $\mathbf{e}_x$ and $\mathbf{e}_y$ are SBERT embeddings of the two texts.
SBERT Similarity is particularly useful for QA evaluation when lexical overlap is low but semantic
meaning is preserved.

\subsection{DeepSeek-V3.2 Backbone}\label{app:experiment:deepseek}

Notably, as shown in Table \ref{tab:deepseek}, when all methods use DeepSeek-V3.2 \cite{deepseek2025deepseek} as the backbone, MemWeaver still achieves the best performance across all question categories, indicating that its gains are not attributable to backbone choice.
Compared with A-Mem, MemWeaver yields substantial improvements on the more reasoning-intensive categories, most notably on \emph{Temporal} questions (F1: 55.52 vs.\ 41.36; BLEU-1: 45.36 vs.\ 31.90; RGE-2: 26.97 vs.\ 15.67), while also improving \emph{Multi-Hop} (F1: 31.35 vs.\ 31.13) and \emph{Single-Hop} (F1: 45.19 vs.\ 41.58). In contrast, MemoryBank performs poorly across all categories (e.g., Temporal F1: 2.43), suggesting that flat retrieval over unstructured memories is insufficient even under a strong backbone, whereas MemWeaver benefits from temporally grounded and structured retrieval.


\subsection{Comparison Results}\label{app:experiment:comp}

Across backbones, MemWeaver consistently improves both exact-match and semantics-aware metrics. 
As shown in Table~\ref{tab:EM}, under GPT-4o-mini MemWeaver achieves the best EM across all four categories 
(e.g., Temporal: 11.21; Single-Hop: 16.77) and ranks first overall for both EM and METEOR (1.00/1.50). 
Similar trends hold for smaller backbones such as Qwen2.5-1.5B, where MemWeaver substantially increases EM on 
Temporal (8.10) and Single-Hop (12.96), again achieving the top overall ranking.

Beyond exact matching, Table~\ref{tab:SBERT} shows that MemWeaver also yields strong gains on sequence-level and 
semantic similarity metrics. In particular, it consistently achieves the highest ROUGE-L and SBERT scores across 
most categories and backbones (e.g., GPT-4o-mini Temporal: ROUGE-L 49.98, SBERT 76.03; Qwen2.5-1.5B Temporal: 
ROUGE-L 44.67, SBERT 72.69), indicating that its answers are not only more precise but also more semantically 
aligned with the references. Together, these results suggest that MemWeaver improves both factual exactness and 
semantic faithfulness across diverse backbone capacities.

\begin{table*}[!htbp]
\centering
\scriptsize
\setlength{\tabcolsep}{3.2pt}
\renewcommand{\arraystretch}{1.05}

\newcommand{\g}[1]{\cellcolor{gray!20}{#1}}
\resizebox{\textwidth}{!}{%
\begin{tabular}{l l|c c c c c c c c|c c}
\toprule
\multirow{3}{*}{\textbf{Model}} & \multirow{3}{*}{\textbf{Method}}
& \multicolumn{8}{c|}{\textbf{Category}} & \multicolumn{2}{c}{\textbf{Overall}} \\
\cmidrule(lr){3-10}\cmidrule(lr){11-12}
& &
\multicolumn{2}{c}{\textbf{Multi-Hop}} &
\multicolumn{2}{c}{\textbf{Temporal}} &
\multicolumn{2}{c}{\textbf{Open-Domain}} &
\multicolumn{2}{c|}{\textbf{Single-Hop}} &
\multicolumn{2}{c}{\textbf{Ranking}} \\
& &
\textbf{EM} & \textbf{METEOR} &
\textbf{EM} & \textbf{METEOR} &
\textbf{EM} & \textbf{METEOR} &
\textbf{EM} & \textbf{METEOR} &
\textbf{EM} & \textbf{METEOR} \\
\midrule

\multirow{5}{*}{\mname{GPT}{-4o-mini}}
& LoCoMo
& 0.35 & \textbf{15.56}
& 0.00 & 9.87
& 2.08 & 7.67
& 6.54 & \textbf{39.75}
& 3.50 & 2.00 \\
& MemoryBank
& 0.00 & 7.57
& 0.00 & 3.95
& 2.08 & 7.26
& 0.00 & 13.04
& 4.25 & 4.25 \\
& ReadAgent
& 0.35 & 5.46
& 0.00 & 4.76
& 0.00 & 3.69
& 0.00 & 8.01
& 4.25 & 4.75 \\
& A-Mem
& 1.77 & 13.74
& 3.74 & 19.35
& 4.17 & 8.65
& 9.27 & 32.79
& 2.00 & 2.50 \\
& \g{\textbf{MemWeaver}}
& \g{\textbf{4.26}} & \g{14.20}
& \g{\textbf{11.21}} & \g{\textbf{25.25}}
& \g{\textbf{6.25}} & \g{\textbf{9.70}}
& \g{\textbf{16.77}} & \g{33.46}
& \g{\textbf{1.00}} & \g{\textbf{1.50}} \\
\midrule

\multirow{5}{*}{\mname{Llama3.2}{-3B}}
& LoCoMo
& 0.71 & 4.71
& 0.00 & 3.17
& 1.04 & 4.57
& 0.71 & 9.47
& 3.25 & 3.25 \\
& MemoryBank
& 0.00 & 6.51
& 0.00 & 2.30
& 0.00 & \textbf{6.39}
& 0.00 & 11.15
& 4.63 & 2.75 \\
& ReadAgent
& 0.00 & 1.21
& 0.62 & 2.33
& 1.04 & 3.39
& 0.00 & 2.46
& 3.63 & 4.75 \\
& A-Mem
& 0.71 & 5.61
& 0.62 & 4.42
& 1.04 & 3.51
& 3.92 & \textbf{16.68}
& 2.50 & 2.50 \\
& \g{\textbf{MemWeaver}}
& \g{\textbf{1.06}} & \g{\textbf{6.60}}
& \g{\textbf{2.49}} & \g{\textbf{5.48}}
& \g{\textbf{2.08}} & \g{3.98}
& \g{\textbf{5.35}} & \g{16.26}
& \g{\textbf{1.00}} & \g{\textbf{1.75}} \\
\midrule

\multirow{5}{*}{\mname{Llama3.2}{-1B}}
& LoCoMo
& \textbf{0.71} & 5.92
& 0.00 & 5.81
& 3.12 & 7.25
& 1.07 & 10.77
& 1.75 & 3.00 \\
& MemoryBank
& 0.00 & 6.06
& 0.00 & 1.76
& 1.04 & 6.66
& 0.00 & 9.27
& 3.00 & 3.75 \\
& ReadAgent
& 0.00 & 2.97
& 0.00 & 1.31
& 1.04 & 7.13
& 1.07 & 5.36
& 2.75 & 4.75 \\
& A-Mem
& 0.00 & 5.97
& \textbf{0.62} & 7.50
& 0.00 & \textbf{7.26}
& 0.00 & 12.33
& 3.00 & 2.00 \\
& \g{\textbf{MemWeaver}}
& \g{0.00} & \g{\textbf{6.11}}
& \g{\textbf{0.62}} & \g{\textbf{8.23}}
& \g{\textbf{4.17}} & \g{7.21}
& \g{\textbf{1.78}} & \g{\textbf{13.14}}
& \g{\textbf{1.25}} & \g{\textbf{1.50}} \\
\midrule

\multirow{5}{*}{\mname{Qwen2.5}{-1.5B}}
& LoCoMo
& 0.35 & 6.01
& 0.00 & 3.74
& 1.04 & 9.44
& 1.07 & 13.48
& 3.25 & 3.25 \\
& MemoryBank
& 0.00 & 7.58
& 0.00 & 3.49
& 0.00 & 7.06
& 0.00 & 14.08
& 4.63 & 3.75 \\
& ReadAgent
& 0.00 & 3.67
& 0.00 & 1.88
& 1.04 & 8.97
& 0.71 & 5.52
& 3.88 & 4.50 \\
& A-Mem
& 0.71 & 8.14
& 1.25 & 7.01
& 1.04 & 7.51
& 3.80 & 20.55
& 2.25 & 2.50 \\
& \g{\textbf{MemWeaver}}
& \g{\textbf{2.84}} & \g{\textbf{11.34}}
& \g{\textbf{8.10}} & \g{\textbf{20.84}}
& \g{\textbf{7.29}} & \g{\textbf{10.99}}
& \g{\textbf{12.96}} & \g{\textbf{25.66}}
& \g{\textbf{1.00}} & \g{\textbf{1.00}} \\
\bottomrule
\end{tabular}%
}
\caption{Experimental results on the LoCoMo dataset across four question types (Multi-Hop, Temporal, Open-Domain, and Single-Hop). Results are reported in EM and METEOR (\%). EM denotes Exact Match. Note that due to the strictness of the EM metric, some methods receive zero scores in certain categories. Best results within each backbone are in bold, and MemWeaver is highlighted in gray. Ranking indicates the average rank across categories (Rank 1 is best; lower is better), computed separately for EM and METEOR.}
\label{tab:EM}
\end{table*}

\begin{table*}[!htbp]
\centering
\scriptsize
\setlength{\tabcolsep}{3.2pt}
\renewcommand{\arraystretch}{1.05}

\newcommand{\g}[1]{\cellcolor{gray!20}{#1}}

\resizebox{\textwidth}{!}{%
\begin{tabular}{l l|c c c c c c c c|c c}
\toprule
\multirow{3}{*}{\textbf{Model}} & \multirow{3}{*}{\textbf{Method}}
& \multicolumn{8}{c|}{\textbf{Category}} & \multicolumn{2}{c}{\textbf{Overall}} \\
\cmidrule(lr){3-10}\cmidrule(lr){11-12}
& &
\multicolumn{2}{c}{\textbf{Multi-Hop}} &
\multicolumn{2}{c}{\textbf{Temporal}} &
\multicolumn{2}{c}{\textbf{Open-Domain}} &
\multicolumn{2}{c|}{\textbf{Single-Hop}} &
\multicolumn{2}{c}{\textbf{Ranking}} \\
& &
\textbf{RGE-L} & \textbf{SBERT} &
\textbf{RGE-L} & \textbf{SBERT} &
\textbf{RGE-L} & \textbf{SBERT} &
\textbf{RGE-L} & \textbf{SBERT} &
\textbf{RGE-L} & \textbf{SBERT} \\
\midrule

\multirow{5}{*}{\mname{GPT}{-4o-mini}}
& LoCoMo
& 24.88 & 45.78
& 23.09 & 40.56
& 16.65 & 40.09
& 39.77 & 51.85
& 2.25 & 2.50 \\
& MemoryBank
& 5.08 & 32.25
& 4.21 & 26.10
& 5.06 & 32.79
& 6.92 & 32.04
& 5.00 & 4.25 \\
& ReadAgent
& 9.45 & 28.67
& 13.12 & 45.07
& 5.76 & 26.72
& 9.92 & 26.78
& 4.00 & 4.50 \\
& A-Mem
& 22.41 & 45.44
& 38.38 & 67.86
& 13.93 & 36.89
& 35.75 & 50.73
& 2.75 & 2.75 \\
& \g{\textbf{MemWeaver}}
& \g{\textbf{25.68}} & \g{\textbf{46.51}}
& \g{\textbf{49.98}} & \g{\textbf{76.03}}
& \g{\textbf{22.44}} & \g{\textbf{42.71}}
& \g{\textbf{40.82}} & \g{\textbf{54.63}}
& \g{\textbf{1.00}} & \g{\textbf{1.00}} \\
\midrule

\multirow{5}{*}{\mname{Llama3.2}{-3B}}
& LoCoMo
& 9.01 & 27.33
& 7.45 & 18.84
& 7.35 & 28.76
& 12.31 & 26.33
& 3.00 & 4.00 \\
& MemoryBank
& 3.85 & 31.53
& 1.52 & 19.65
& 3.08 & \textbf{32.10}
& 5.29 & 31.68
& 4.50 & 2.50 \\
& ReadAgent
& 1.78 & 17.40
& 3.01 & 12.02
& 5.22 & 19.63
& 2.51 & 14.63
& 4.50 & 5.00 \\
& A-Mem
& 19.05 & 38.51
& 15.43 & 30.57
& 7.73 & 30.47
& 28.90 & 41.72
& 2.00 & \textbf{1.75} \\
& \g{\textbf{MemWeaver}}
& \g{\textbf{20.22}} & \g{\textbf{38.82}}
& \g{\textbf{19.41}} & \g{\textbf{34.53}}
& \g{\textbf{11.37}} & \g{29.35}
& \g{\textbf{34.26}} & \g{\textbf{44.61}}
& \g{\textbf{1.00}} & \g{\textbf{1.75}} \\
\midrule

\multirow{5}{*}{\mname{Llama3.2}{-1B}}
& LoCoMo
& 11.06 & 30.65
& 15.54 & 46.56
& 13.23 & 37.84
& 14.63 & 31.55
& 3.25 & 3.25 \\
& MemoryBank
& 3.68 & 29.62
& 1.58 & 17.25
& 4.51 & 31.30
& 5.20 & 26.39
& 5.00 & 4.75 \\
& ReadAgent
& 6.49 & 29.26
& 4.62 & 26.45
& 14.29 & 39.19
& 8.03 & 26.44
& 3.50 & 4.00 \\
& A-Mem
& 11.53 & \textbf{35.17}
& 17.35 & 50.99
& 13.74 & \textbf{42.88}
& 20.29 & 34.46
& 2.25 & \textbf{1.50} \\
& \g{\textbf{MemWeaver}}
& \g{\textbf{13.99}} & \g{34.08}
& \g{\textbf{26.71}} & \g{\textbf{58.69}}
& \g{\textbf{15.17}} & \g{40.60}
& \g{\textbf{23.92}} & \g{\textbf{36.83}}
& \g{\textbf{1.00}} & \g{\textbf{1.50}} \\
\midrule

\multirow{5}{*}{\mname{Qwen2.5}{-1.5B}}
& LoCoMo
& 10.42 & 29.68
& 7.45 & 25.59
& 12.12 & 35.66
& 12.70 & 30.14
& 3.00 & 3.75 \\
& MemoryBank
& 5.70 & 31.33
& 3.02 & 18.83
& 4.75 & 31.22
& 8.11 & 31.84
& 4.50 & 4.00 \\
& ReadAgent
& 7.14 & 28.20
& 2.81 & 27.27
& 12.63 & 35.13
& 7.88 & 26.33
& 4.00 & 4.25 \\
& A-Mem
& 14.69 & 36.91
& 24.32 & 60.23
& 10.93 & 37.22
& 21.63 & 38.44
& 2.50 & 2.00 \\
& \g{\textbf{MemWeaver}}
& \g{\textbf{21.29}} & \g{\textbf{47.66}}
& \g{\textbf{44.67}} & \g{\textbf{72.69}}
& \g{\textbf{18.81}} & \g{\textbf{43.41}}
& \g{\textbf{34.04}} & \g{\textbf{48.67}}
& \g{\textbf{1.00}} & \g{\textbf{1.00}} \\
\bottomrule
\end{tabular}%
}
\caption{Experimental results on the LoCoMo dataset across four question types (Multi-Hop, Temporal, Open-Domain, and Single-Hop). Results are reported in RGE-L and SBERT (\%). RGE-L denotes ROUGE-L. Best results within each backbone are in bold, and MemWeaver is highlighted in gray. Ranking indicates the average rank across categories (Rank 1 is best; lower is better), computed separately for RGE-L and SBERT.}
\label{tab:SBERT}
\end{table*}

\subsection{Results on the Adversarial Dataset}\label{app:adv}




\begin{table}[t]
\centering
\small
\setlength{\tabcolsep}{6pt}
\renewcommand{\arraystretch}{1.12}
\newcommand{\g}[1]{\cellcolor{gray!20}{#1}}

\begin{tabular}{l|c c c}
\toprule
\textbf{Metric} 
& \textbf{MemoryBank} 
& \textbf{A-Mem} 
& \g{\textbf{MemWeaver}} \\
\midrule
EM        & 26.68 & 22.42 & \g{\textbf{37.00}} \\
F1        & 29.13 & 25.14 & \g{\textbf{72.20}} \\
ROUGE-2  & 24.78 & 20.88 & \g{\textbf{70.83}} \\
ROUGE-L  & 29.03 & 25.39 & \g{\textbf{72.22}} \\
BLEU-1   & 28.46 & 24.36 & \g{\textbf{64.61}} \\
\bottomrule
\end{tabular}

\caption{Results on the LoCoMo \textbf{Adversarial} category using GPT-4o mini. All scores are reported as percentages (\%). Best results are in bold, and MemWeaver is highlighted in gray.}
\label{tab:adversarial}
\end{table}

On the LoCoMo adversarial task, MemWeaver consistently outperforms MemoryBank and A-Mem across all metrics with GPT-4o mini. In particular, MemWeaver achieves over 10\% gains in EM and large improvements in F1, ROUGE-2, and ROUGE-L, indicating stronger robustness under adversarial perturbations. These results suggest that MemWeaver is more resilient to misleading or conflicting memory signals, preserving both factual correctness and semantic alignment. Overall, the consistent improvements across exact-match and generation-based metrics validate the effectiveness of MemWeaver’s structured memory consolidation in adversarial settings.

\subsection{Case Study}\label{app:experiment:case}

\begin{table*}[t]
\centering
\small
\setlength{\tabcolsep}{6pt}
\renewcommand{\arraystretch}{1.2}
\begin{tabularx}{\textwidth}{@{}p{0.12\textwidth} p{0.28\textwidth} X X X@{}}
\toprule
\textbf{Category} & \textbf{Question} & \textbf{A-Mem} & \textbf{MemWeaver (Ours)} & \textbf{Reference} \\
\midrule

\textbf{Multi-Hop}
& What Jon thinks the ideal dance studio should look like?
& A place where people can express themselves through dance.
& \textbf{By the water.}
& \textcolor{red}{By the water}, with natural light and Marley flooring. \\
\midrule
\textbf{Temporal}
& When is Jon's group performing at a festival?
& Next month.
& \textbf{February, 2023.}
& \textcolor{red}{February, 2023.} \\
\midrule
\textbf{Open-Domain}
& What would Caroline's political leaning likely be?
& LGBTQ activist group, passionate about rights and community support.
& \textbf{Likely liberal.}
& \textcolor{red}{Liberal.} \\
\midrule
\textbf{Single-Hop}
& What kind of flooring is Jon looking for in his dance studio?
& The context does not specify what kind of flooring Jon is looking for in his dance studio.
& \textbf{Marley flooring.}
& \textcolor{red}{Marley flooring.} \\

\bottomrule
\end{tabularx}
\caption{Case study examples comparing A-Mem and MemWeaver across four question types.}
\label{tab:case-study}
\end{table*}

To qualitatively compare MemWeaver with A-Mem, we present four representative
examples from different types of questions. These examples highlight three key strengths of MemWeaver.

\textbf{(i) Multi-Hop compositional reasoning:} MemWeaver leverages GM-based compositional retrieval to bridge evidence scattered across multiple sessions, constructing implicit relational chains that support Multi-Hop reasoning (e.g., linking clues to pinpoint ``by the water'' as the defining studio feature).

\textbf{(ii) Temporal grounding:} By explicitly storing normalized time information in memory, MemWeaver can return absolute and grounded timestamps (e.g., ``February, 2023'') rather than relative expressions, which is essential for accurate temporal reasoning.

\textbf{(iii) High-precision memory retrieval:} MemWeaver improves the precision of stored memory by jointly leveraging structured triples and traceable passages, preserving complete and correct evidence. This allows it to provide accurate supporting context for both Single-Hop factual queries and open-domain questions.

Overall, these cases illustrate that MemWeaver consistently produces
answers that are more specific, temporally grounded, and closely aligned with
the underlying evidence.
By integrating structured compositional reasoning with evidence-linked memory
retrieval, MemWeaver is able to recover salient attributes, normalize
temporal information, and output concise conclusions that directly match the
question intent.
This combination enables more reliable long-term conversational question
answering across diverse reasoning types.

\subsection{Model Scaling Analysis}\label{app:experiment:scale}

\begin{figure*}[!htbp]
    \centering
    \begin{minipage}[t]{0.48\textwidth}
        \centering
        \includegraphics[width=\linewidth]{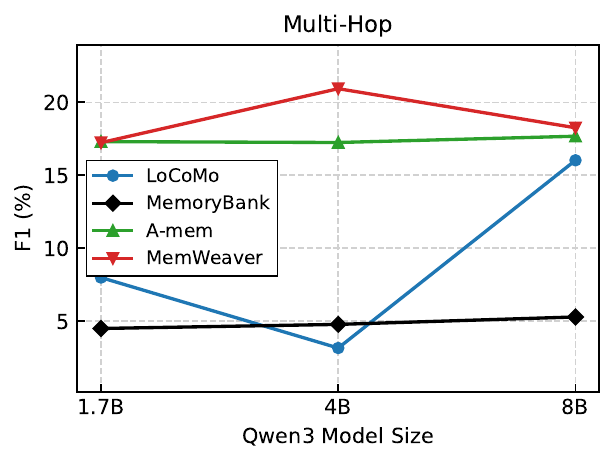}
        \vspace{-0.3em}
        {\small (a) Multi-Hop}
        \label{fig:qwen3_multihop}
    \end{minipage}\hfill
    \begin{minipage}[t]{0.48\textwidth}
        \centering
        \includegraphics[width=\linewidth]{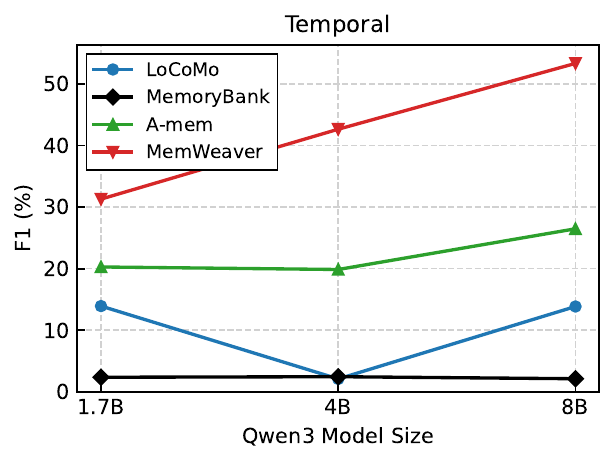}
        \vspace{-0.3em}
        {\small (b) Temporal}
        \label{fig:qwen3_temporal}
    \end{minipage}

    \vspace{0.8em} 

    \begin{minipage}[t]{0.48\textwidth}
        \centering
        \includegraphics[width=\linewidth]{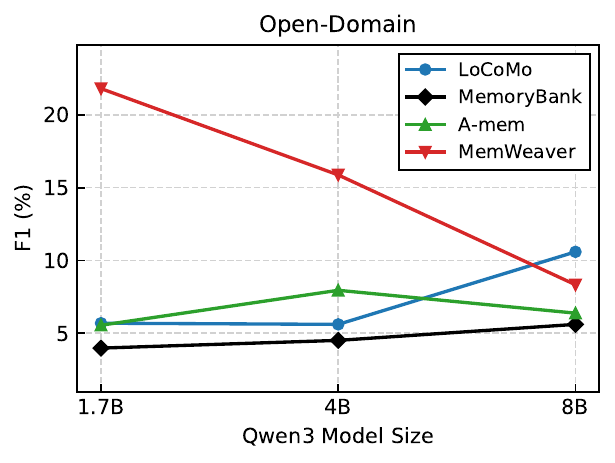}
        \vspace{-0.3em}
        {\small (c) Open-Domain}
        \label{fig:qwen3_opendomain}
    \end{minipage}\hfill
    \begin{minipage}[t]{0.48\textwidth}
        \centering
        \includegraphics[width=\linewidth]{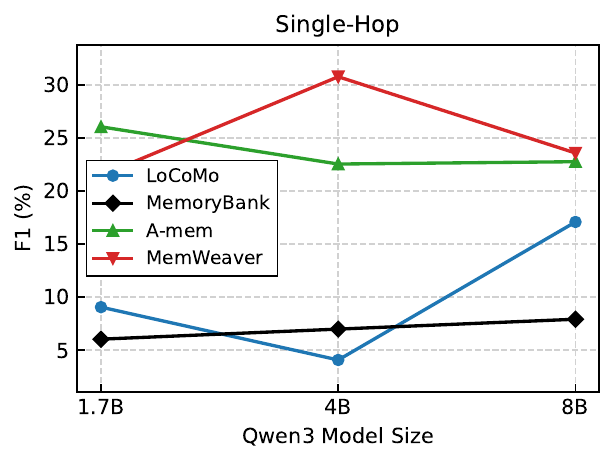}
        \vspace{-0.3em}
        {\small (d) Single-Hop}
        \label{fig:qwen3_singlehop}
    \end{minipage}

    \caption{
    Model scaling results on Qwen3-1.7B, 4B, and 8B backbones.
    Each panel reports the F1 performance across backbone scales for one LoCoMo question category,
    comparing LoCoMo, MemoryBank, A-Mem, and MemWeaver.
    Lines indicate how each method's category-level F1 changes as the backbone model size increases.
    }
    \label{fig:qwen3_scaling}
\end{figure*}

Figure~\ref{fig:qwen3_scaling} shows that MemWeaver generally outperforms the baselines across Qwen3 backbone scales on all question categories. 
Overall, MemWeaver demonstrates stable and competitive performance as the backbone model size varies.

\subsection{Human Evaluation Details}\label{app:experiment:human}
We conduct a human evaluation to assess the quality of retrieved knowledge, focusing on whether the retrieved evidence is sufficient to answer the question correctly.
Following prior work, we randomly sample 25 questions per category (Multi-Hop, Temporal, Open-Domain, and Single-Hop) from the LoCoMo dataset.

For each question, annotators are provided with:
\begin{itemize}
    \item The question,
    \item The knowledge retrieved by the memory system, and
    \item The ground-truth reference answer.
\end{itemize}

We recruit five NLP experts who are Master’s or PhD students actively working in the NLP field, with research experience in retrieval-augmented generation and LLM agents.
Each question is independently annotated by all five experts.
Annotators are asked to make a binary judgment (\emph{helpful} / \emph{not helpful}) indicating whether the retrieved knowledge contains sufficient information to derive the reference answer.
The annotators come from diverse geographic regions, with three based in Asia and two based in Oceania.
We report the average helpfulness rate across the five annotators. Annotators are compensated at a rate of \$0.10 per annotation, for a total of \$50 in human evaluation costs.

\onecolumn
\section{Algorithm}\label{app:algorithm}
\subsection{Memory Construction}\label{app:algorithm-cons}

\begin{algorithm}[H]
\caption{KG Writing and Maintenance}
\label{alg:kg-build}
\begin{algorithmic}[1]
\Require New dialogue unit $x_i=\langle q_i,a_i,s_i,t_i\rangle$, KG $G$
\Ensure Updated KG $G$ and triple index $\mathcal{I}_T$
\State Create passage node $p_i$ with text and metadata $(s_i,t_i)$
\State $\mathcal{V}\leftarrow \mathrm{LLM}(x_i \mid P_{\text{ent}})$ \Comment{entities}
\State $\mathcal{R}\leftarrow \mathrm{LLM}(x_i,\mathcal{V} \mid P_{\text{rel}})$ \Comment{candidate triples}
\ForAll{$r=\langle h,\rho,u\rangle \in \mathcal{R}$}
  \State $\hat{t}\leftarrow \eta(x_i,r)$; attach metadata $m(r)$
  \State Insert/merge entity nodes; add semantic edge $r$ into $G$
\EndFor
\State Link $p_i$ to involved entities via structural edges
\State $\Omega\leftarrow \mathrm{LLM}(G,\text{session} \mid P_{\text{review}})$
\State Apply $\Omega$; remove redundant relations; rebuild $\mathcal{I}_T$
\end{algorithmic}
\end{algorithm}

\clearpage

\begin{algorithm}[H]
\caption{Experience Induction and Online Update}
\label{alg:exp-update}
\begin{algorithmic}[1]
\Require New dialogue unit $x_i$; clusters $\{C_j\}$ with centers $\{\mu_j\}$; pending buffer $\mathcal{B}$;
thresholds $(\tau_{\mathrm{high}},\tau_{\mathrm{low}})$; candidate shortlist size $K$; update trigger $B_{\mathrm{add}}$; recluster window $B_{\mathrm{re}}$
\Ensure Updated clusters $\{C_j\}$, centers $\{\mu_j\}$, and experience items $\mathcal{E}$

\State $e_i \leftarrow \phi(\mathrm{text}(x_i))$
\State $j^\star \leftarrow \arg\max_j \cos(e_i,\mu_j)$;
$s^\star \leftarrow \max_j \cos(e_i,\mu_j)$

\If{$s^\star \ge \tau_{\mathrm{high}}$}
  \State $C_{j^\star} \leftarrow C_{j^\star} \cup \{x_i\}$
  \State $\mathtt{add\_buffer}(C_{j^\star}) \leftarrow \mathtt{add\_buffer}(C_{j^\star}) \cup \{x_i\}$
\ElsIf{$\tau_{\mathrm{low}} \le s^\star < \tau_{\mathrm{high}}$}
  \State $\mathcal{J}(x_i) \leftarrow \operatorname{Top}_{K}\!\bigl(\{\mu_j\}, e_i\bigr)$
  \State $\hat{j} \leftarrow
  \mathrm{LLM}\!\left(
  x_i,\ \{(\mathtt{center\_text}_j,\mathcal{S}_j)\}_{j\in\mathcal{J}(x_i)}
  \mid P_{\mathrm{route}}
  \right)$
  \If{$\hat{j} \neq \texttt{none}$}
    \State $C_{\hat{j}} \leftarrow C_{\hat{j}} \cup \{x_i\}$
    \State $\mathtt{add\_buffer}(C_{\hat{j}}) \leftarrow \mathtt{add\_buffer}(C_{\hat{j}}) \cup \{x_i\}$
  \Else
    \State $\mathcal{B} \leftarrow \mathcal{B} \cup \{x_i\}$
  \EndIf
\Else
  \State $\mathcal{B} \leftarrow \mathcal{B} \cup \{x_i\}$
\EndIf

\ForAll{clusters $C_j$ with $|\mathtt{add\_buffer}(C_j)| \ge B_{\mathrm{add}}$}
  \State $\mu_j \leftarrow \frac{1}{|C_j|}\sum_{x\in C_j}\phi(\mathrm{text}(x))$
  \State $\mathtt{center\_text}_j \leftarrow \mathrm{LLM}\!\left(C_j \mid P_{\mathrm{sum}}\right)$
  \State $\mathcal{E}_j \leftarrow \mathrm{LLM}\!\left(C_j \mid P_{\mathrm{ind}}\right)$
  \State $\mathcal{E}_j \leftarrow \operatorname{Filter}(\mathcal{E}_j)$
  \State clear $\mathtt{add\_buffer}(C_j)$
\EndFor

\If{$|\mathcal{B}| \ge B_{\mathrm{re}}$}
  \State $\{C'_k\} \leftarrow \operatorname{DBSCAN}(\mathcal{B})$; clear $\mathcal{B}$
  \ForAll{candidate clusters $C'_k$}
    \State $b_k \leftarrow \mathrm{LLM}\!\left(C'_k \mid P_{\mathrm{coh}}\right)$
    \If{$b_k = \texttt{yes}$}
      \State $C_{\mathrm{new}} \leftarrow C'_k$
      \State $\mu_{\mathrm{new}} \leftarrow \frac{1}{|C_{\mathrm{new}}|}\sum_{x\in C_{\mathrm{new}}}\phi(\mathrm{text}(x))$
      \State $\mathtt{center\_text}_{\mathrm{new}} \leftarrow \mathrm{LLM}\!\left(C_{\mathrm{new}} \mid P_{\mathrm{sum}}\right)$
      \State $\mathcal{E}_{\mathrm{new}} \leftarrow \mathrm{LLM}\!\left(C_{\mathrm{new}} \mid P_{\mathrm{ind}}\right)$
      \State $\mathcal{E}_{\mathrm{new}} \leftarrow \operatorname{Filter}(\mathcal{E}_{\mathrm{new}})$
    \Else
      \State $\mathcal{B} \leftarrow \mathcal{B} \cup C'_k$
    \EndIf
  \EndFor
\EndIf
\end{algorithmic}
\end{algorithm}

\subsection{Retrieval and Reasoning}\label{app:algorithm-retrie}

\begin{algorithm}[H]
\caption{Inference Retrieval and Context Assembly}
\label{alg:retrieve}
\begin{algorithmic}[1]
\Require Query $Q$; triple index $\mathcal{I}_T$; graph $G$ (with attached passage/experience nodes); Passage Memory $P$;
budgets $(k_r,k_p,k_e)$
\Ensure $(C_{\mathrm{KG}}, C_{\mathrm{TXT}})$
\State $\mathcal{R}_{\mathrm{seed}} \leftarrow
\operatorname{Retrieve}( \mathcal{I}_T, Q, k_r )$
\State $\mathcal{R}_{\mathrm{cand}} \leftarrow
\operatorname{Expand}( G, \mathcal{R}_{\mathrm{seed}}, 1 )$
\State $\mathcal{R}_{\mathrm{cand}} \leftarrow
\operatorname{Filter}( \mathcal{R}_{\mathrm{cand}}, Q )$
\State $\mathcal{R}_{\mathrm{llm}} \leftarrow
\mathrm{LLM}( Q, \mathcal{R}_{\mathrm{cand}} \mid P_{\mathrm{select}} )$
\State $\mathcal{R}^{\star} \leftarrow
\mathcal{R}_{\mathrm{llm}} \cup
\operatorname{Top}_{k_r}( \mathcal{R}_{\mathrm{cand}}; Q )$
\State $\mathcal{R}^{\star} \leftarrow
\operatorname{Deduplicate}( \mathcal{R}^{\star} )$
\State $\mathcal{P}_{\mathrm{kg}} \leftarrow
\operatorname{Collect}( G, \mathcal{R}^{\star}, \text{passage} )$
\State $\mathcal{E}_{\mathrm{kg}} \leftarrow
\operatorname{Collect}( G, \mathcal{R}^{\star}, \text{experience} )$
\State $\mathcal{P}_{\mathrm{glob}} \leftarrow
\operatorname{Retrieve}( P, Q, k_p )$
\State $\mathcal{P}^{\star} \leftarrow
\operatorname{RankDedup}( \mathcal{P}_{\mathrm{kg}} \cup \mathcal{P}_{\mathrm{glob}}, Q )$
\State $\mathcal{E}^{\star} \leftarrow
\operatorname{RankDedup}( \mathcal{E}_{\mathrm{kg}}, Q )$
\State $C_{\mathrm{KG}} \leftarrow
\operatorname{Serialize}( \mathcal{R}^{\star} )$
\State $C_{\mathrm{TXT}} \leftarrow
\operatorname{Assemble}( \mathcal{P}^{\star}_{1:k_p}, \mathcal{E}^{\star}_{1:k_e} )$
\Return $(C_{\mathrm{KG}}, C_{\mathrm{TXT}})$
\end{algorithmic}
\end{algorithm}

\end{document}